%% file: main.tex
\ifdefined\useorstyle

\documentclass[opre,nonblindrev]{informs3}

\usepackage{macros/packages-or}
\usepackage{macros/editing-macros}
\usepackage{macros/statistics-macros-or}
\usepackage{macros/notation}
\usepackage{macros/formatting}

\DoubleSpacedXI %

\usepackage{endnotes}
\let\footnote=\endnote

\usepackage{natbib}
 \bibpunct[, ]{(}{)}{,}{a}{}{,}%
 \def\bibfont{\small}%
 \def\bibsep{\smallskipamount}%

\TheoremsNumberedThrough     %
\ECRepeatTheorems

\EquationsNumberedThrough    %

\usepackage{hyperref}

\usepackage{pgfplotstable}
\usepackage{graphicx}
\usepackage{subcaption}
\usepackage{float} 
\usepackage{tabularx}
\usepackage{overpic}
\usepackage{tikz}
\usepackage{rotating}
\usepackage{psfrag}
\usepackage{enumitem}
  
\begin{document}

\RUNAUTHOR{Cai, Namkoong, and Yadlowsky}
\RUNTITLE{Diagnosing Model Performance Under Distribution Shift}

\TITLE{Diagnosing Model Performance Under Distribution Shift}

\ARTICLEAUTHORS{%

  \AUTHOR{Tiffany (Tianhui) Cai}
\AFF{Department of Statistics, Columbia University, New York, NY 10027, \EMAIL{tiffany.cai@columbia.edu}} %

\AUTHOR{Hongseok Namkoong}
\AFF{Decision, Risk, and Operations Division, Columbia Business School, New York, NY 10027, \EMAIL{namkoong@gsb.columbia.edu}} %

\AUTHOR{Steve Yadlowsky}
\AFF{Google Research, Brain Team, Mountain View, CA 94043, \EMAIL{yadlowsky@google.com}} %

} %

\ABSTRACT{\input{abstract}}

\KEYWORDS{distribution shift, machine learning, reliability}

\maketitle

\else

\documentclass[11pt]{article}
\usepackage[numbers]{natbib}
\usepackage{macros/packages}
\usepackage{macros/editing-macros}
\usepackage{macros/formatting}
\usepackage{macros/statistics-macros}
\usepackage{macros/notation}

\input{notation}
\begin{document}

\abovedisplayskip=8pt plus0pt minus3pt
\belowdisplayskip=8pt plus0pt minus3pt

\begin{center}
  {\huge  Diagnosing Model Performance Under Distribution Shift} \\
  \vspace{.5cm}
  {\Large Tiffany (Tianhui) Cai$^{1}$  ~~~ Hongseok Namkoong$^{2}$ ~~~
    Steve Yadlowsky$^{3}$} \\
  \vspace{.2cm}
  {\large Columbia University$^{1, 2}$ \hspace{3em} Google Research$^{3}$ \\
  Department of Statistics$^1$,
  Decision, Risk, and Operations Division$^2$, Brain Team$^3$} \\
  \vspace{.2cm}
  \texttt{tiffany.cai@columbia.edu, namkoong@gsb.columbia.edu, yadlowsky@google.com}
\end{center}

\begin{abstract}%
  \input{abstract}
\end{abstract}

\fi

\input{introduction}

\input{decomp}

\input{estimation}

\input{experiments}

\input{theory}

\input{discussion}

\input{ack}

\bibliographystyle{abbrvnat}

\ifdefined\useorstyle
\setlength{\bibsep}{.0em}
\else
\setlength{\bibsep}{.7em}
\fi

\bibliography{../macros/bib}

\ifdefined\useorstyle

\ECSwitch

\ECHead{Appendix}

\else
\newpage
\appendix

\fi

\input{theory_proofs}

\ifdefined\short
\section{Generalization to other decompositions}
    \input{decomp_gen}
\fi
\input{more_estimation}

\input{experiment_details}

\input{alternative_s}

\end{document}

%% file: abstract.tex
Prediction models can perform poorly when deployed to target distributions
different from the training distribution.  To understand these operational
failure modes, we develop a method, called DIstribution Shift DEcomposition
(DISDE), to attribute a drop in performance to different types of distribution
shifts. Our approach decomposes the performance drop into terms for 1) an
increase in harder but frequently seen examples from training, 2) changes in
the relationship between features and outcomes, and 3) poor performance on
examples infrequent or unseen during training.
Empirically, we demonstrate how our method can 1) inform potential modeling
improvements across distribution shifts for employment prediction on tabular
census data, and 
and 2) help to explain why certain domain adaptation
methods fail to improve model performance for satellite image classification.

%% file: notation.tex
\newcommand{\Appendix}{Appendix}
\newcommand{\z}{w}
\renewcommand{\Z}{W}

%% file: introduction.tex
\section{Introduction}
\label{section:intro}

Prediction models operate on data distributions different from those seen
during training, but often perform worse on these \emph{target} distributions
than on the original \emph{training} distribution.  For example,
\citet{WongEtAl21} observed EPIC's sepsis risk assessment model, which is
deployed across hundreds of hospitals in the US, performs ``substantially
worse'' in the wild compared to vendor claims.  The lack of reliability in
predictive performance across distributions has been documented in many
domains including healthcare~\cite{LeekEtAl10, BandiEtAl18, ZechBaLiCoTiOe18,
  ChenPiRoJoFeGh20, WongEtAl21}, loan approval~\cite{Hand06}, wildlife
conservation~\cite{BeeryCoGj20}, and education~\cite{AmorimCaVe18}.
Similar performance degradation has been widely observed in computer vision
and natural language processing (NLP) settings~\cite{RechtRoScSh19,
  MillerKrReSc20, TaoriDaShCaReSc20,
  ShankarDaRoRaReSc19,MillerTaRaSaKoShLiCaSc21,LazaridouKuGrAgLiTe21}.
As decisions are increasingly made based on model predictions, we need
rigorous and scalable tools for diagnosing model failures in order to guide
resources toward effective model improvements.

Different distribution shifts require different solutions. Understanding
\emph{why} model performance worsened is a fundamental step for informing
subsequent methodological and operational interventions.  While there are
multiple taxonomies for discussing distribution shifts \cite{LiptonWaSmo18,
  ScholkopfJaPeSgZhMo12, TranLiDuPhCoRe22}, we focus on a popular
one: %
we categorize distribution shifts as either a change in the marginal
distribution of the covariates ($X$), or a change in the conditional
relationship between the label and covariate ($Y \mid X$).  Since real
distribution shifts occur as a combination of both types, we develop a
framework for understanding how much performance degradation is assigned to
each type in order to best improve model performance.

There are many reasons why the marginal distribution of the covariates $X$ can
change.  For example, training and target data may be collected from different
points in time and space, marginalized groups may be underrepresented in the
training data, or population demographics may change over
time~\cite{Shimodaira00, Ben-DavidBlCrPe07, ChenLaDaPaKe14,
  BuolamwiniGe18}. If the shift in the distribution of covariates $X$ is
small, then domain adaptation~\cite{Shimodaira00},
warm-starting/fine-tuning~\cite{ColomboGoGr11, WortsmanIlLiKiHaFaNaSc21}, or
importance sampling-based reweighting~\cite{AsmussenGl07, Owen15, DuchiHaNa22}
approaches can be effective.  If there is a more extreme form of covariate
shift, such as the presence of target covariate values $X$ that were unseen
during training, then it may be necessary to collect data and labels over this
new group.

On the other hand, shifts in the relationship between the outcomes and
covariates $Y\mid X$ can be caused by changes in measurement errors, changes
in user behavior, and unrecorded confounding factors whose distributions
change~\cite{Hand06}.  In these cases, recent works on explicit causal or
worst-case modeling may be helpful \cite{XuCaMa12, GaoChKl17,
  BlanchetKaZhMu17, RothenhauslerBuMePe18, BertsimasGuKa18, KuhnEsNgSh19,
  VanParysEsKu21, DuchiNa21}. When methods like these cannot address the
performance drop, existing data from the training distribution may not be
useful for the target distribution, so that further data collection and
labeling may be necessary. Alternatively, the covariates $X$ may not be
predictive of the outcome between both distributions, requiring changes in
feature engineering.

\begin{figure}[t]
  \vspace{-1cm}
  \centering
  \hspace{2cm}
  \includegraphics[scale=.4]{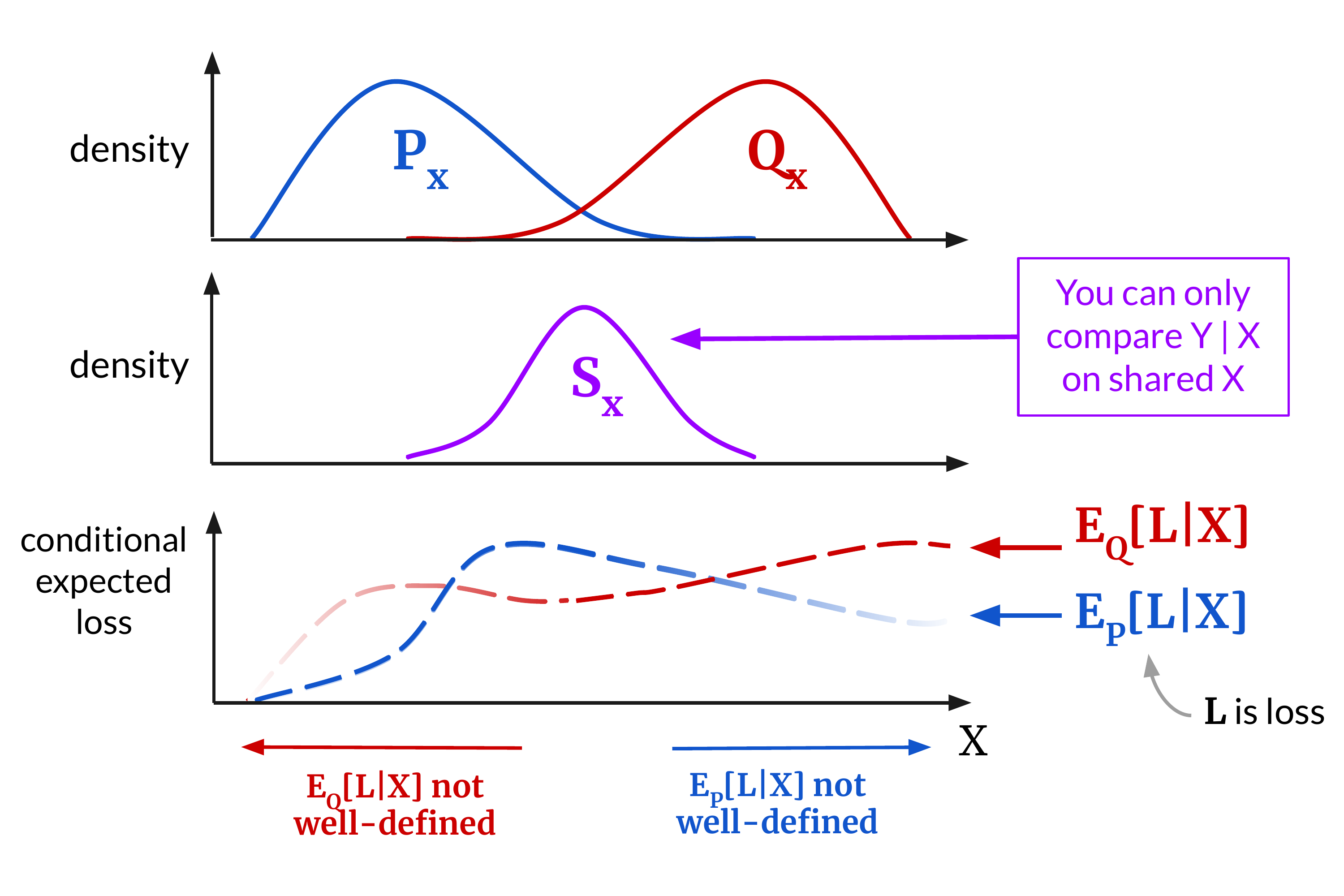}
  \caption{\label{fig:intro} \textbf{Diagnosing distribution shifts}:
    Illustration of a key idea on a one-dimensional
    covariate $X$: 
    we can only compare
    $Y \mid X$ for training vs test, and predictive performances thereof, 
    on $X$'s that are common
    between both distributions.
    We use $L$ to denote the loss. 
    }
\end{figure}

Towards diagnosing model performance degradations, we
propose a new framework which we call
DIstribution Shift DEcomposition (DISDE).
At a high level,
performance degradation attributed to $Y\mid X$ shift is obtained
by varying $Y\mid X$ between training and test while keeping the distribution 
of $X$ fixed,
and performance degradation attributed to $X$ shift is obtained by
varying the distribution of $X$ while keeping $Y\mid X$ fixed. 
In order to make these comparisons, observe that
the
training and target
distributions of $Y\mid X$---and predictive performances thereof---can only be
compared over values of $X$ that are sufficiently common between training and target
(see Figure~\ref{fig:intro}).
Thus, one core aspect of our framework is the notion of a 
\emph{shared distribution} between $X$'s of
the training and target distributions over which it is easy to 
estimate model performace (Section \ref{sec:estimation}).
Using this shared distribution, 
we decompose the performance degradation into three terms: one for changes in the
relationship between $X$ and $Y$ on this shared distribution, and two for changes in the distribution of $X$ from both the training and target distributions to the shared distribution. 

By attributing loss to different
distribution shifts, our approach goes beyond detecting changes in the
distribution, as DISDE quantifies how much each type of shift affects
performance. The first term in our decomposition (Equation~\eqref{eq:decomp} to come)
measures performance degradation due to an increase in the frequency of harder
values of $X$ that are common between training and target distributions (an
$X$ shift).  The second term measures performance degradation due to changes in the
relationship between $X$ and $Y$ (a $Y\mid X$ shift), over values of $X$ common to both
distributions.  The third measures performance degradation due to poor performance on
values of $X$ that are infrequent or unseen during training (an $X$
shift). 

One important aspect of DISDE is precisely defining and estimating the shared
distribution.  As we discuss in Section~\ref{section:shared}, 
we define this shared distribution 
so that it has the largest mass on values of $X$ that are most likely in both
training and target.  Our approach is motivated by weighting schemes in causal
inference that focus on a subset of the population where treatment and control
units are similar enough to compare~\cite{CrumpHoImMi06, LiMoZa18}.  To
estimate this shared distribution, we train a binary classifier to predict
between the training and target distributions using features $X$, which we
refer to as the \emph{auxiliary domain classifier}.  Using this classifier, we
can reweight samples from training and test into the shared distribution; we
operationalize ``common'' examples across train and target as those that
cannot be confidently and correctly classified as being from either training
or target.

Our approach is general, and DISDE can be extended to attribute performance
degradation to use $Z$ in place of $X$ for \emph{any} flexible feature vector
$Z$ (\ifdefined\short\Appendix\else Section\fi
~\ref{section:general-decomp}). For example, $Z$ can be a subset of covariates
(Section \ref{sec:adult_missing}, a (learned) feature mapping (Section
\ref{sec:wilds}), some other piece of metadata, or even the label $Y$ itself,
turning DISDE into a method for diagnosing label shift.

  We apply our method to three
real distribution shifts in Section \ref{sec:experiments}.  In the first
example (Section \ref{sec:adult}), we predict employment from tabular census
data~\cite{DingHaMoSc21} and study how predictive performance deteriorates
over a variety of natural and semi-synthetic distribution shifts. We
illustrate how DISDE gives correct performance attributions, and how it can
inform different ways of improving performance on the target distribution.  
In
the second example (Section \ref{sec:wilds}), we study satellite image
classification for land or building use (FMoW-WILDS~\cite{KohSaEtAl20}).  We
use DISDE to diagnose why a popular domain adaptation method,
DANN~\cite{GaninUsAjGeLaLaMaLe16}, does not do better than standard empirical
risk minimization over spatiotemporal shifts, even though DANN is designed to
perform well over distribution shifts.  This example also demonstrates how
DISDE can be scaled to high-dimensional, complex data such as images.
Overall, our experiments show how DISDE can be used to guide modeling
improvements and to provide insight where specific interventions fail, 
and also to understand learned feature representations.

Finally, we analyze the large-sample statistical properties of the proposed
estimation method in Section~\ref{sec:theory}. By leveraging results from the
semiparametric statistics literature~\cite{NeweyMc94}, we give a central limit
theorem for our estimator even when the auxiliary domain classifier is
estimated nonparametrically.  This provides a principled approach for
generating confidence intervals in DISDE. We compare the asymptotic variance
of our estimator to lower bounds under a well-developed calculus for
statistical functionals~\cite{BickelKlRiWe98, Newey94, Kennedy22} to confirm
that our proposed approach is statistically efficient, under suitable
regularity assumptions on the auxiliary classifier.

To summarize, our main contributions are as follows:
\begin{itemize}
    \item We introduce a simple method, DIstribution Shift DEcomposition
        (DISDE), to attribute a change in model performance across a
        distribution shift 
        to $X$ and $Y\mid X$ shifts.
    \item We empirically demonstrate how DISDE can inform model improvements
        across distribution shifts 
        on both tabular and image data.
    \item We analyze the asymptotic properties of DISDE, provide a principled
        way to estimate confidence intervals, and show efficiency of our
        estimator under certain conditions.
\end{itemize}
As there is a large body of work on distribution shifts across multiple
fields, we defer a discussion of related literature to
Section~\ref{section:discussion} where we situate the current work in a
broader context.

%% file: decomp.tex
\section{Decomposing the performance degradation}
\label{sec:decomp}

Consider a model $f$ that predicts outcome $Y$ from covariates $X$.
Let $f$ be trained on data
$(X,Y)$ 
drawn from the \emph{training} distribution $P$. Let $\loss(f(x), y)$ 
be a loss
function denoting a notion of predictive error (e.g. 0-1 error, cross-entropy
loss).  We consider a situation in which we observe a performance degradation
from $P$ to a \emph{target} distribution $Q$,
i.e. $\E_Q[\loss(f(X),Y)]>\E_P[\loss(f(X),Y)]$. 
As a motivating example,
consider a model trained to predict employment status based on demographic
information, which could be important for economic policy-making and
marketing.
\setlength{\columnsep}{0pt}%
\setlength{\intextsep}{10pt}%
\begin{wraptable}{r}{0.35\textwidth}
    \centering
    \begin{tabular}{r|l}
        Training accuracy & 70\% \\
        Target accuracy & 61\% \\
    \end{tabular}
    \vspace{0.5em}
    \caption{Performance degradation in the employment
    prediction example}
    \label{tab:bars}
\end{wraptable}
Due to operational constraints, the model may be trained using data
collected from one state, such as West Virginia (training distribution),
but used across a range
of different states, including Maryland (target distribution). In such cases, 
we may only have enough labeled data from Maryland to evaluate a model, but not
enough to train a new model. 
As we show in
Table~\ref{tab:bars} and further investigate in Section~\ref{sec:adult}, 
the model trained on West Virginia
performs poorly on Maryland. As collecting more training data from Maryland is
costly, we seek to understand the performance drop to determine how best to
improve our models for use on Maryland.

To understand how the distribution shift led to such a performance
degradation, we attribute the performance degradation to
shifts in the marginal distribution of $X$ and to the conditional
distribution of $Y \mid X$. The natural objects to quantify $X$ and $Y\mid X$
shifts are, respectively, the marginal distributions of $X$, $P_{X}(\cdot)$
and $Q_{X}(\cdot)$, and the conditional risks on $P$ and $Q$,
\begin{equation*}
  R_{\mu}(x) := \E_\mu[\ell(f(X),Y)\mid X = x]~\mbox{for}~\mu=P, Q.
\end{equation*}
However, there are two practical issues with studying these quantities
directly. First, estimation of these functions is a challenging density
estimation or nonparametric regression problem. As in our motivating example
above, one of our primary use cases is when we only have limited labeled
target data from $Q$.
Second, even if we could derive reasonable estimates of these functions, it
would be difficult to display them in a way that would be useful for a human
to understand, especially for $X$ of larger dimension. Therefore, we need to
summarize the infinite dimensional quantities
$\{R_P(\cdot), R_Q(\cdot), P_X, Q_X\}$ in a practical and statistically
efficient way.

Our approach is the following: to attribute loss to $X$ shift, compare the
conditional risk $R_P(x)$ (resp. $R_Q(x)$) under different marginal
distributions of $X$. Then, to attribute loss to $Y\mid X$ shift, compare
$R_P(x)$ and $R_Q(x)$ under the same marginal distribution of $X$.  Recalling
Figure~\ref{fig:intro}, the main challenge in doing this is that $R_P(x)$
(resp. $R_Q(x)$) is only well-defined for values of $x$ that are in the
support of $P_X$ (resp. $Q_X$). Therefore, when comparing performance
$R_Q(x) - R_P(x)$ across different $Y\mid X$ distributions, we need to define a
\emph{shared distribution} $\S_{X}$ with support contained in both $P_X$ and
$Q_X$.
We temporarily defer a discussion
on different choices of formulations for shared distributions to
Section~\ref{section:shared} and first outline our approach for any fixed
definition of the shared distribution $S_X$.

\begin{figure}[h]
    \vspace{1em}
        \centering
        \input{rectangles}
        \vspace{1em}
        \caption{Diagram of decomposition: left to right represents changing $X$
    distribution from $P_X$ to $S_X$ to $Q_X$. Up to down represents changing
    $Y\mid X$
    from $P_{Y|X}$ to $Q_{Y|X}$, by replacing $R_P(X)$ with $R_Q(X)$. 
    The bolded arrows 
    correspond to the decomposition terms we propose, where each arrow
    $A\to B$ represents the difference $B-A$.
    }
    \label{fig:rectangles}
\end{figure}
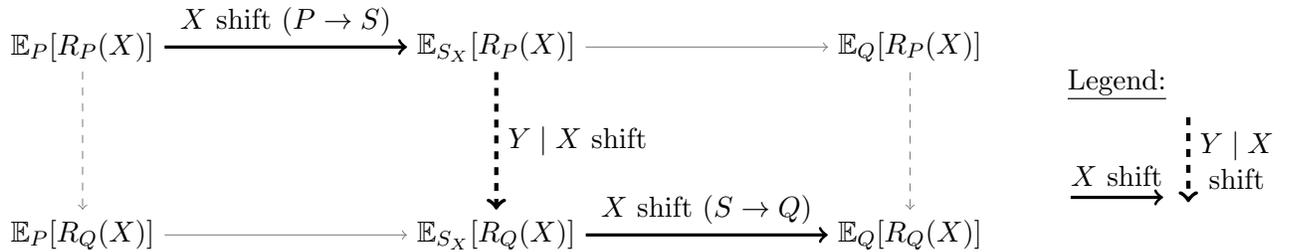

We thus attribute performance degradation to $Y\mid X$ by comparing 
the averages of the conditional risks $R_P(X)$ and $R_Q(X)$ on the
marginal distribution $S_{X}$ as
\begin{equation*}
    \E_{S_X}[ R_{\bm{P}}(X) ]~\mbox{versus}~\E_{S_X}[ R_{\bm{Q}}(X) ]
.\end{equation*}
Then, for $X$ shift, we can construct a well-defined comparison of the average of $R_P(X)$
under the two marginal distributions $P_X$ and $S_{X}$,
\begin{equation*}
    \E_{\bm{P_X}}[ R_P(X) ]~\mbox{versus}~\E_{\bm{S_X}}[ R_P(X) ],
\end{equation*}
and similarly for the average of $R_Q(X)$ under the marginal distributions $Q_X$
and $S_{X}$.  Altogether, we can decompose the change in expected loss from
$P$ to $Q$ as the telescoping sum
\begin{subequations}
  \label{eq:decomp}
  \begin{align}
    \E_Q[\ell(f(X),Y)]-\E_P[\ell(f(X),Y)] 
      & = \E_{\S_X}[R_P(X)]-\E_{P}[R_P(X)] & \text{$X$ shift ($P\rightarrow S$)} \\
    &\hspace{1.5em}
      + \E_{\S_X}[R_Q(X)-R_P(X)] & \text{$Y\mid X$ shift}  \\
    &\hspace{1.5em}
      + \E_{Q}[R_Q(X)]-\E_{\S_X}[R_Q(X)].  & \text{$X$ shift ($S\rightarrow Q$)}
  \end{align}
\end{subequations}
The telescoping sum is illustrated in Figure
\ref{fig:rectangles}, where the bolded arrows represent the terms in
Equation~\eqref{eq:decomp}.

\subsection{Interpreting the decomposition~\eqref{eq:decomp}}
\label{section:interpretation}

We provide intuition for the terms in Equation \eqref{eq:decomp}
and discuss ways to improve test performance for when each term is large.

\subsubsection{$X$ shift ($P\to S$) term}
$\E_{\S_X}[R_P(X)]-\E_{P_X}[R_P(X)]$ is the expected difference in loss under
$P_{Y|X}$ between examples with $X\sim P_X$ and $X\sim \S_X$. 
This term is large if the model performs worse on
covariates $X$ common in both $P_X$ and $Q_X$ under $P_{Y|X}$, compared to
covariates from $P_X$.  This can happen if e.g. $Q_X$ has more
of the hard examples from $P_X$. 
If $P_{Y|X}=Q_{Y|X}$, then we expect the shared data $X\sim \S_X$ to be able
to inform good models under $Q_X$, e.g. by using domain adaptation methods. 
\ifdefined\short
\else

In the context of the employment prediction problem described at the beginning
of the section, a large $X$ shift ($P\to S$) term could indicate that the change in loss between West
Virginia and Maryland is largely due to Maryland having more people than
West Virginia of
demographics that were difficult for the model to predict. 
In this case, 
we may refit the model by reweighting training
observations from West Virginia to resemble the population in Maryland.
\fi

\subsubsection{$Y\mid X$ shift term} %
$\E_{\S_X}[R_Q(X)-R_P(X)]$ is the expected
difference in loss between $Q_{Y|X}$ and $P_{Y|X}$ over the shared covariates
$X\sim S_X$ common across $P_X$ and $Q_X$.  We expect this term to be large if
the loss is larger under $Q_{Y|X}$ than under $P_{Y|X}$ over shared covariates
$X\sim \S_X$.  Loosely speaking, this can happen if the relationship $Y\mid X$
changes unfavorably from $P$ to $Q$ for the prediction model $f$. 
In cases with a large $Y\mid X$ shift term, it may be
necessary to collect and label new data over $Q_{Y|X}$, or to modify the set of
covariates. 
\ifdefined\short
\else

In the context of employment prediction, a large $X$ shift ($S\to Q$) term may indicate
that the change in model loss is largely due to a difference in the likelihood
of whether a person \emph{of the same demographics}
is employed, in West Virginia vs Maryland, for people whose demographics
are common in both states. 
Solving this may require investing in a new
data collection campaign in Maryland large enough to train a new 
model. Alternatively, we may search for new covariates $X'$ such that $Y\mid (X, X')$
remains similar across West Virginia (training) and Maryland (target) and
train a new model that also uses the additional features $X'$.
\fi

\subsubsection{$X$ shift ($S\to Q$) term} %
$\E_{Q_X}[R_Q(X)]-\E_{\S_X}[R_Q(X)]$ is the
expected difference in loss under $Q_{Y|X}$ between examples $X\sim \S_X$ and
$X\sim Q_X$.  Consider values of $X$ that are frequent in the target
distribution $Q_X$ but infrequent or unseen in the training distribution
$P_X$. We colloquially refer to these as ``\new examples'', which include 
unseen examples as well as values of $X$ that are relatively more common in
$Q_X$ compared to $\S_X$.
The $X$ shift ($S\to Q$) term is large if the model performs poorly
incurs higher loss on such ``\new examples'' under $Q_{Y|X}$.  
This type of performance degradation can be addressed by collecting ``\new
examples'' and retraining or fine-tuning a model on these examples.%
\ifdefined\short
\else

In our employment prediction example, 
a large $X$ shift ($S\to Q$) term could indiciate
that the change in model loss is largely due to Maryland having people of
demographics that are unseen or uncommon West Virginia, that are also difficult for
the model to predict. 
A potential solution involves conducting a survey of
employment status in Maryland targeting demographic groups that were rarely
observed in West Virginia. 
\fi
\subsubsection{Comparisons}
The two $X$ shift terms both attribute performance
degradation to $X$ shift, but differ in the following way: the $X$ shift ($P\to
S$) term contains performance degradation
from $P_X$ having support on examples with low loss where $Q_X$ does not, and
the $X$ shift ($S\to Q$) term contains performance
degradation from $Q_X$ having support on examples with high loss where $P_X$
does not. 
However, beyond this distinction, how much a value of $X$ is ``allocated'' to 
either of these terms may not necessarily be meaningful.
Another difference between the two $X$ shift terms 
is that the $X$ shift ($P\to S$) term
uses $P_{Y|X}$ while the $X$ shift ($S\to Q$) term uses $Q_{Y|X}$. 
Because of this, one part of the $X$ shift (from
$P_X$ to $S_X$) is measured under
$P_{Y|X}$, while the rest (fom $S_X$ to $Q_X$) is measured under $Q_{Y|X}$. 
Again, this distinction 
can be somewhat arbitrary. %
Despite these subtleties, in practice, it may be 
helpful to think of these two terms as qualitatively similar
when diagnosing
a model's failures in practice.
\tc{except one of our examples totally does this}

For 
concrete examples of decompositions for
different $Y\mid X$ vs $X$ shifts
and 
the corresponding modeling interventions,
please refer to the experiments in Section \ref{sec:adult}. 

\subsection{Defining the shared distribution $S_X$}
\label{section:shared}

To operationalize our approach, we choose a specific shared distribution $S_X$
over $X$ whose support is contained in that of $P_X$ and $Q_X$. An ideal
choice of shared distribution is one that has higher density when the
$P_X$ and $Q_X$ both have higher density, and lower density
when either of $P_X$ or $Q_X$ has a low density. This sharpens the above
requirement of shared support to a stronger notion of sharing regions of high
density. Letting $p_{X}$, $q_{X}$, and $s_{X}$ be the densities of $X$ under
$P$, $Q$, and $S$
under a suitable base measure, consider choosing
\begin{equation}
    \label{eq:s}
    \s_X(x) \propto \frac{p_X(x)q_X(x)}{p_X(x)+q_X(x)}.
\end{equation}
However, there are many shared distributions that satisfy our criteria, and
the above choice is not unique. For example, other choices include
\begin{subequations}
  \label{eq:alternative_s}
  \begin{align}
    \s_X(x)
    & \propto \min\{ p_X(x), q_X(x) \}
      ~~~\mbox{or} ~~~ \label{eq:s-minpq} \\
    \s_X(x)
    & \propto
      \begin{cases}
        p_X(x) + q_X(x)  ~~
        & \mbox{if}~
        \min\left\{\frac{p_X(x)}{q_X(x)}, \frac{q_X(x)}{p_X(x)}\right\} \ge \epsilon \\
        0 ~~ & \mbox{otherwise}
      \end{cases}
    \label{eq:s-truncation}
\end{align}
\end{subequations}
for some $\epsilon > 0$. The shared distribution~\eqref{eq:s-truncation} is
particularly intuitive as it defines ``shared'' examples as those that have
sufficiently high likelihood ratios. However, it is difficult to
operationalize as it depends on an arbitrary choice of $\epsilon > 0$.
Observe that for all of the above definitions of $S_X$, if $P_X=Q_X$,
then $S_X=P_X=Q_X$. 
Additionally observe that 
Equations~\eqref{eq:s} and \eqref{eq:s-minpq} become similar
(up to a scaling factor) for regions of $x$ where
$p_X(x)\gg q_X(x)$ or $p_X(x)\ll q_X(x)$.

For the shared distributions listed above, our decomposition can be estimated
by reweighting data from $P$ and $Q$ to approximate samples from the
shared distribution, as described in more detail in Section \ref{sec:estimation}. 
We use an auxiliary binary classifier between $P$ and $Q$
to learn these importance weights, which are bounded under weak assumptions
and provide low (asymptotic) variance estimates for $\E_{S_X}[R_P(X)]$
and $\E_{\S_X}[R_Q(X)]$. Most of our method
extends to the alternative shared distributions~\eqref{eq:alternative_s} in place
of the one in Equation~\eqref{eq:s}, and in practice we find that the qualitative
conclusions obtained from our approach are not very sensitive to the specific
choice of shared distribution 
(\Appendix~\ref{sec:alternative_s}). For ease of exposition, we focus on the
shared distribution defined in Equation~\eqref{eq:s}.
\ifdefined\short
See \Appendix~\ref{section:general-decomp} for a generalization of this
decomposition to arbitrary mappings $(X,Y)\mapsto \genfeature$ in place of $X$. 
\else
\subsection{Generalization to other decompositions}
    \input{decomp_gen}

\fi

%% file: rectangles.tex
\begin{tikzpicture}[node distance=4cm and 2cm, auto]
  \node (eprp) at (0,0) {$\E_P[R_P(X)]$};
  \node (eprq) at (0,-2.5) {$\E_P[R_Q(X)]$};
  \node (esrp) at (5.5,0) {$\E_{S_X}[R_P(X)]$};
  \node (esrq) at (5.5,-2.5) {$\E_{S_X}[R_Q(X)]$};
  \node (eqrp) at (11,0) {$\E_Q[R_P(X)]$};
  \node (eqrq) at (11,-2.5) {$\E_Q[R_Q(X)]$};
  \node (lr_l) at (13, -2) {};
  \node (lr_r) at (14.5, -2) {};
  \node (ud_d) at (14.7, -0.8) {};
  \node (ud_u) at (14.7, -2.2) {};
    \node (legend) at (13.75, -0.5) {\underline{Legend:} };
    \draw[->, very thick, solid] (eprp) to node {$X$ shift ($P\to S$)} (esrp);
    \draw[->, very thick, solid] (esrq) to node {$X$ shift ($S\to Q$)} (eqrq);
  \draw[->, very thick, solid] (lr_l) to node {$X$ shift} (lr_r);
    \draw[->, ultra thick, dashed] (esrp) to node {$Y\mid X$ shift} (esrq);
   \draw[->, ultra thick, dashed] (ud_d) to node[align=center] [right] {$Y\mid
    X$\\ shift} (ud_u);
   \draw[->, gray, dashed] (eprp) to node {} (eprq); 
   \draw[->, gray, dashed] (eqrp) to node {} (eqrq); 
   \draw[->, gray] (eprq) to node {} (esrq); 
   \draw[->, gray] (esrp) to node {} (eqrp); 
\end{tikzpicture}

%% file: decomp_gen.tex
\label{section:general-decomp}

Our approach generalizes to any choice of mapping
$(X,Y) \mapsto \genfeature$ in place of $X$. This generalization attributes
performance degradation from $P$ to $Q$ to changes in the marginal distributions
of $\genfeature$, and to changes
in the conditional distribution of the data $(X, Y)$ given
$\genfeature$. Specifically, we can generalize $R_\mu(\genfeatureval)$ to be
\begin{equation*}
  R_{\mu}(\genfeatureval) := \E_\mu[\ell(f(X),Y)|\genfeature=\genfeatureval]~\mbox{for}~\mu=P, Q,
\end{equation*}
and re-write the decomposition as
\begin{subequations}
  \label{eq:decomp-gen}
  \begin{align}
    \E_Q[\ell(f(X),Y)]-\E_P[\ell(f(X),Y)]  
    & = \E_{\S_\genfeaturesub}[R_P(\genfeature)]-\E_{P}[R_P(\genfeature)] &
      \text{$Z$ shift ($P\to S$)} \\
    &\hspace{1.5em}
      + \E_{\S_\genfeaturesub}[R_Q(\genfeature)-R_P(\genfeature)] &
      \text{$(X,Y)\mid Z$ shift}  \\
    &\hspace{1.5em}
      + \E_{Q}[R_Q(\genfeature)]-\E_{\S_\genfeaturesub}[R_Q(\genfeature)]. &
      \text{$Z$ shift ($S\to Q$)}
  \end{align}
\end{subequations}

This generalization allows us to provide a richer diagnostic for machine
learning applications.  For example, 
when $Z = \phi(X)$ where
$\phi(\cdot)$ is a feature mapping (e.g. from the prediction model $f$), 
large performance
degradation attributed to $Y\mid \phi(X)$ shift may imply
the relationship between $Y$ and $\phi(X)$ is not stable between training and
target. This could suggest that those features are not appropriate 
for prediction across both distributions.  This is demonstrated in experiments in Section
\ref{sec:adult_missing}, where $\genfeature$ is a subset of $X$, and Section
\ref{sec:wilds}, where $\genfeature$ is a learned feature representation.

We can also use metadata as $Z$. In this case, we can think of $X$
as containing both metadata and ``regular'' data, 
and $f(\cdot)$ to only use regular data. 
As an additional example, we can use $Y$ as $Z$, in which case the decomposition
measures label shift
\cite{LiptonWaSmo18}.

The methodological (Section \ref{sec:estimation}) and theoretical (Section
\ref{sec:theory}) results in the rest of the paper 
extend immediately
to using $Z$ instead of $X$; we would replace $\pi(x), R_P(x), R_Q(x)$
with $\pi(z), R_P(z), R_Q(z)$ and similarly for their estimates.

%% file: estimation.tex
\section{Estimation}
\label{sec:estimation}

In our decomposition~\eqref{eq:decomp}, each term corresponds to the difference of
consecutive pairs of $\E_{P}[R_P(\feature)]$,
$\E_{\S_\featuresub}[R_P(\feature)]$, $\E_{\S_\featuresub}[R_Q(\feature)]$,
$\E_{Q}[R_Q(\feature)]$, as illustrated in Figure \ref{fig:rectangles}.  
The first and last terms are easily estimated by the
empirical average of losses over $(X_i, Y_i) \simiid P$ and $(X_j,
Y_j) \simiid Q$, respectively.
Thus, we focus on the estimation of the middle two terms
\begin{equation}
  \label{eq:two-terms}
  \E_{\S_\featuresub}[R_P(\feature)]~~~\mbox{and}~~~\E_{\S_\featuresub}[R_Q(\feature)].
\end{equation}
A key statistical challenge is that we do not have access to samples from the
shared distribution $\S_{\featuresub}$ since it is a fictitious quantity
defined for interpretating the performance degradation. As we only have
samples $\feature_i \simiid P$ and $ \feature_j \simiid Q$, our strategy is to reweight
them to approximate samples from $S_{\featuresub}$, which in turn allows us to
estimate the two terms~\eqref{eq:two-terms}.
To see this, rewrite our estimands as\begin{subequations}
  \label{eq:importance-sampling}
\begin{align}
  \theta_P \defeq \E_{\S_\featuresub}[R_P(\feature)]
  & =\E_{P}\left[R_P(\feature)
    \frac{d\S_\featuresub}{dP_\featuresub}(\feature)\right]
      =\E_{P}\left[\ell(f(X), Y)
    \frac{d\S_\featuresub}{dP_\featuresub}(\feature)\right]
    \label{eq:importance-p} \\
  \theta_Q \defeq  \E_{\S_\featuresub}[R_Q(\feature)]
  & =\E_{Q}\left[R_Q(\feature)
    \frac{d\S_\featuresub}{dQ_\featuresub}(\feature)\right]
     =\E_{Q}\left[\ell(f(X),Y)
    \frac{d\S_\featuresub}{dQ_\featuresub}(\feature)\right].
    \label{eq:importance-q}
\end{align}
\end{subequations}
Once we obtain estimates for the importance weights $\frac{dS_X}{dP_X}$ and
$\frac{dS_X}{dQ_X}$, we estimate $\theta_P$ and $\theta_Q$ by using an
empirical plug-in of Equation \eqref{eq:importance-sampling} above, and 
as described in
Algorithm~\ref{alg:algo}.

To calculate the importance weights 
$\frac{dS_X}{dP_X},\; \frac{dS_X}{dQ_X}$, we can use  
an auxiliary domain classifier. %
Recalling the
definition of the shared distribution $S_\featuresub$~\eqref{eq:s}, we have
\begin{align}
  \label{eq:importance-ratio}
        \frac{dS_\featuresub}{dP_\featuresub}(\featureval) \propto \frac{q(\featureval)}{p(\featureval)+q(\featureval)}~~~\mbox{and}~~~
  \frac{dS_\featuresub}{dQ_\featuresub}(\featureval) \propto \frac{p(\featureval)}{p(\featureval)+q(\featureval)}.
\end{align}

Now observe that $\frac{dS_\featuresub}{dP_\featuresub}\propto
\frac{q(\featureval)}{p(\featureval)+q(\featureval)}$ corresponds to the
probability that a data point with value $x$
drawn from an even mixture distribution of $P_\featuresub$ and $Q_\featuresub$ actually came from
$Q_\featuresub$,
which can be estimated using any machine learning classifier that outputs class
probabilities (e.g. neural networks, random forests, logistic regression), 
as we can consider coming from $Q_\featuresub$ vs $P_\featuresub$ to be the classes for
classification. 
We can thus leverage 
a broad set of scalable and effective machine learning methods
to estimate these probabilities.

Observe also that
weighting samples from $P_\featuresub$ by the classifier probability that it came from
$Q_\featuresub$ is consistent with the intuition that data points likely to be from the
shared distribution $S_\featuresub$ are those that are difficult to classify correctly as
being from $P_\featuresub$ or $Q_\featuresub$.

In practice, we may not have an even mixture of samples from $P_\featuresub$ and $Q_\featuresub$ 
on which to learn the auxiliary domain classifier, e.g. if we have fewer samples
from target than training.
In such cases, we can use standard techniques from the label-shift literature to adjust the base rate \citep{Elkan01}. %
To help formalize this, define the following:
\begin{align}
    \begin{split}
  & \alpha^*~~\mbox{is the proportion of the pooled data that comes from}~ Q_\featuresub \\
  & T = \begin{cases}
    0 ~~&\mbox{if}~\tilde{\feature}~\mbox{is from}~P_\featuresub \\
    1 ~~&\mbox{if}~\tilde{\feature}~\mbox{is from}~Q_\featuresub
    \end{cases}\\
    & \pi^*(\featureval) \defeq \P(T=1\mid\tilde{\feature}=\featureval)
  =\frac{\alpha^* q(\featureval)}{\alpha^* q(\featureval)+(1-\alpha^*) p(\featureval)}
  \label{eq:estimation_m}
    \end{split}
\end{align}
where the last equality is by Bayes' rule, and $\P(\cdot)$ is probability under
the distribution of the observed data.
As before, $\pi^*(\featureval)$ can also be interpreted as the probability output from the true domain classifier.
Then we can express the importance ratios as follows:\footnote{An easy way to see this is that
  $\frac{dS_\featuresub}{dQ_\featuresub}(\featureval),\frac{dS_\featuresub}{dP_\featuresub}(\featureval),
  \pi^*(\featureval)$ can all be written in
  terms of $q(\featureval)/p(\featureval)$.}
\begin{subequations}
    \label{eq:ws}
    \begin{align}
    \frac{dS_\featuresub}{dP_\featuresub}(\featureval) &\propto
        \frac{\pi(\featureval)}{(1-\alpha)\pi(\featureval)+\alpha(1-\pi(\featureval))}
        =:w_P(\pi(\featureval),\alpha)
    \\\frac{dS_\featuresub}{dQ_\featuresub}(\featureval) &\propto
        \frac{1-\pi(\featureval)}{(1-\alpha)\pi(\featureval)+\alpha(1-\pi(\featureval))}
        =:w_Q(\pi(\featureval),\alpha)
    .\end{align}
\end{subequations}
\input{algo}

Since $w_Q(\pi^*(\featureval),\alpha^*)\propto \frac{dS_\featuresub}{dQ_\featuresub}$ and
$w_P(\pi^*(\featureval),\alpha^*)\propto \frac{dS_\featuresub}{dP_\featuresub}$,
we rewrite 
the estimands $\theta_P$ and
$\theta_Q$
from~\eqref{eq:importance-sampling}
as functionals of the observed data distributions $P$ and $Q$, the conditional
probability $\pi^*(\featureval)$, and the mixture proportion $\alpha^*$:
\begin{proposition}
  \label{prop:ipw}
    With $w_P, w_Q$ as defined in Equation~\eqref{eq:ws},
    \begin{equation}
      \label{eqn:ipw}
      \theta_P = \frac{\E_P\left[\ell(f(X), Y) w_P(\pi^*(\feature),\alpha^*) \right]
      }{\E_P\left[w_P(\pi^*(\feature),\alpha^*)\right]}
      ~~~\mbox{and}~~~
      \theta_Q = \frac{\E_Q\left[\ell(f(X), Y) w_Q(\pi^*(\feature),\alpha^*) \right]
      }{\E_Q\left[w_Q(\pi^*(\feature),\alpha^*)\right]}.
\end{equation}
\end{proposition}

The reformulation~\eqref{eqn:ipw} lends itself well to a two-stage
semiparametric estimation technique. In the first stage, we can use various ML
methods or nonparametric statistical estimators to estimate $\pi^*(\featureval)$
as $\what{\pi}(\featureval)$, and an empirical mean to estimate $\alpha^*$ as
$\what{\alpha}$. Then, we plug the estimated quantities into the
\emph{empirical} expectations
\begin{equation}
  \label{eq:estimate_theta_q}
  \what{\theta}_P = 
    \frac{\frac{1}{n_P} \sum_{i=1}^{n_P} \ell(f(X_i), Y_i) w_P(\what{\pi}(\feature_i),\what\alpha)}
    {\frac{1}{n_P} \sum_{i=1}^{n_Q} w_P(\what\pi(\feature_i), \what\alpha)}
~~\mbox{and}~~
\what{\theta}_Q = 
    \frac{\frac{1}{n_Q} \sum_{j=1}^{n_Q} \ell(f(X_j), Y_j) w_Q(\what{\pi}(\feature_j),\what\alpha)}
    {\frac{1}{n_Q} \sum_{i=1}^{n_Q} w_Q(\what{\pi}(\feature_j), \what{\alpha})}.
\end{equation}
See Algorithm~\ref{alg:algo} for a summary of the procedure, 
\Appendix~\ref{sec:more_estimation} for discussion of data splits
and additional implementation details, and 
Section \ref{sec:theory} for statistical properties of the estimators.

Note that the domain classifier probability $\pi^*(\featureval)$ is analogous to the
propensity score in causal inference. Our method is 
motivated by weighting schemes in causal inference that
focus on a subset of the population where treatment and control units are
similar enough to compare~\cite{CrumpHoImMi06, LiMoZa18}.

%% file: algo.tex
\begin{algorithm}[t]
  \caption{ \textsc{Decompose change in loss under shift from $P$ to $Q$}
  \label{alg:algo}
    }
  Estimate $\E_{P}[R_P(\feature)]$ and $\E_{Q}[R_Q(\feature)]$
  using data $(X_i,Y_i)\sim P$, with $i=1,\ldots,n_P$, and
    $(X_j,Y_j)\sim Q$, with $j=1,\ldots,n_Q$:
  \begin{align*}
    \E_{P}[R_P(\feature)] \approx \frac{1}{n_P}\sum_{i=1}^{n_P}
    \ell(f(X_i),Y_i)
    ~~\mbox{and}~~\E_{Q}[R_Q(\feature)] \approx \frac{1}{n_Q}\sum_{j=1}^{n_Q}
    \ell(f(X_j),Y_j).
  \end{align*}

  Estimate $\what{\alpha}=n_Q/(n_P+n_Q)$. 
  
  Estimate $\what{\pi}(\featureval) \approx \P(T=1|\feature=\featureval)$ by
    training a classifier on $\feature$'s
    from $P_\featuresub$ vs $Q_\featuresub$, where samples from $P_\featuresub$
    and $Q_\featuresub$ are given labels 0 and 1, respectively.

  Calculate importance weights proportional to 
  $\frac{dS_\featuresub}{dP_\featuresub}$
  and $\frac{dS_\featuresub}{dQ_\featuresub}$: $$ %
    w_P(\what{\pi}(\featureval), \what{\alpha})
    =\frac{\what{\pi}(\featureval)}{(1-\what{\alpha})\what{\pi}(\featureval)+\what{\alpha}(1-\what{\pi}(\featureval))}
    ~~\mbox{and}~~w_Q(\what{\pi}(\featureval), \what{\alpha})
    =\frac{1-\what{\pi}(\featureval)}{(1-\what{\alpha})\what{\pi}(\featureval)+\what{\alpha}(1-\what{\pi}(\featureval))}.
 $$ %

  Estimate $\E_{\S_\featuresub}[R_P(\feature)]$ and $\E_{\S_\featuresub}[R_Q(\feature)]$ using these
    importance weights: 
  \begin{subequations}
  \begin{align*}
    \E_{\S_\featuresub}[R_P(\feature)]
    &\approx  \frac{\sum_{i=1}^{n_P} \ell(f(X_i), Y_i)
      w_P(\what{\pi}(\feature_i), \what{\alpha})}{
      \sum_{i=1}^{n_P}     w_P(\what{\pi}(\feature_i), \what{\alpha}) } \\
    \E_{\S_\featuresub}[R_Q(\feature)]
    &\approx  \frac{\sum_{j=1}^{n_Q} \ell(f(X_j), Y_j)
      w_Q(\what{\pi}(\feature_j), \what{\alpha})}
      {\sum_{j=1}^{n_Q}w_Q(\what{\pi}(\feature_j), \what{\alpha}) }.
  \end{align*}
  \end{subequations}

    Estimate the terms in Equation~\eqref{eq:decomp} by taking differences between consecutive
  pairs of estimates for
  \begin{align}
    \label{eq:four}
    \E_{P}[R_P(\feature)], \;
    \E_{\S_\featuresub}[R_P(\feature)], \;
    \E_{\S_\featuresub}[R_Q(\feature)],\;
    \E_{Q}[R_Q(\feature)].
  \end{align}
\end{algorithm}

%% file: experiments.tex
\section{Experiments}
\label{sec:experiments}

We demonstrate the versatility of our approach with two sets of experiments.
In the first experiment setting (Section~\ref{sec:adult}), we predict
employment status using tabular census data, as in our motivating example.  We
consider various known distribution shifts and verify that our decomposition
attributes the observed performance degradation to the appropriate types of
shift.  We also illustrate how DISDE can help guide the allocation of
resources toward more effective modeling interventions.  

In the second experiment
setting (Section~\ref{sec:wilds}), we study a satellite image
classification problem where the baseline model performs poorly under
spatiotemporal distribution shifts. We use DISDE to understand why a popular
domain adaptation method (DANN) meant to improve performance under
distribution shift fails to do so.  Our experiments thus demonstrate how DISDE
can be applied in various stages of the model development process 
and also help us understand learned feature representations. 
Code for our method can be found at \mbox{\url{https://github.com/namkoong-lab/disde}}.

\subsection{Case studies: predicting employment from census data}
\label{sec:adult}

In the following case studies, we %
consider natural and semi-synthetic
$Y\mid X$ and $X$ shifts on tabular datasets.
When some interventions are more expensive than
others (e.g., collecting and labeling large quantities of new data vs model
changes), it is
useful to understand which modeling interventions are most likely to help. By
studying specific types of distribution shifts, we illustrate how our
diagnostic can guide modeling interventions such as reweighting existing data,
using domain knowledge to identify and sample a missing covariate, and
collecting new data samples from the target distribution.
Additionally, understanding the distribution shifts that resulted in 
the change in model performance can be useful for its own sake. 

Our task is to predict whether an individual is employed ($Y$) based
on their (tabular) census data ($X$). We use the Adult census
dataset~\cite{DingHaMoSc21} to train an employment classifier on one set of
data (training), and apply the classifier to a different set (target).  We
use the decomposition~\eqref{eq:decomp} and focus on data
from 2018. Both the employment model $f(x)$ and domain classifier
$\hat \pi(x)$ are implemented as random forest classifiers from the
\texttt{sklearn} Python package~\cite{PedregosaVaGrMiThGrBlPr11}. To measure
performance degradation due to distribution shift and not to overfitting, 
all evaluated performances are on validation sets, i.e.,
data that are separate from the training set but are drawn from the same
distribution as the training set. 
Additional implementation details are in
\Appendix~\ref{appendix:adult_details}.
Confidence intervals are in Table~\ref{tab:adult_sd} in
Section~\ref{sec:theory}.

\subsubsection{$Y\mid X$ shift: missing/unobserved covariates}
\label{sec:adult_missing}

\setlength{\columnsep}{5pt}
\begin{wrapfigure}{R}{0.45\textwidth}
\vspace{-0.5em}
            \centering
        \includegraphics[width=\linewidth]{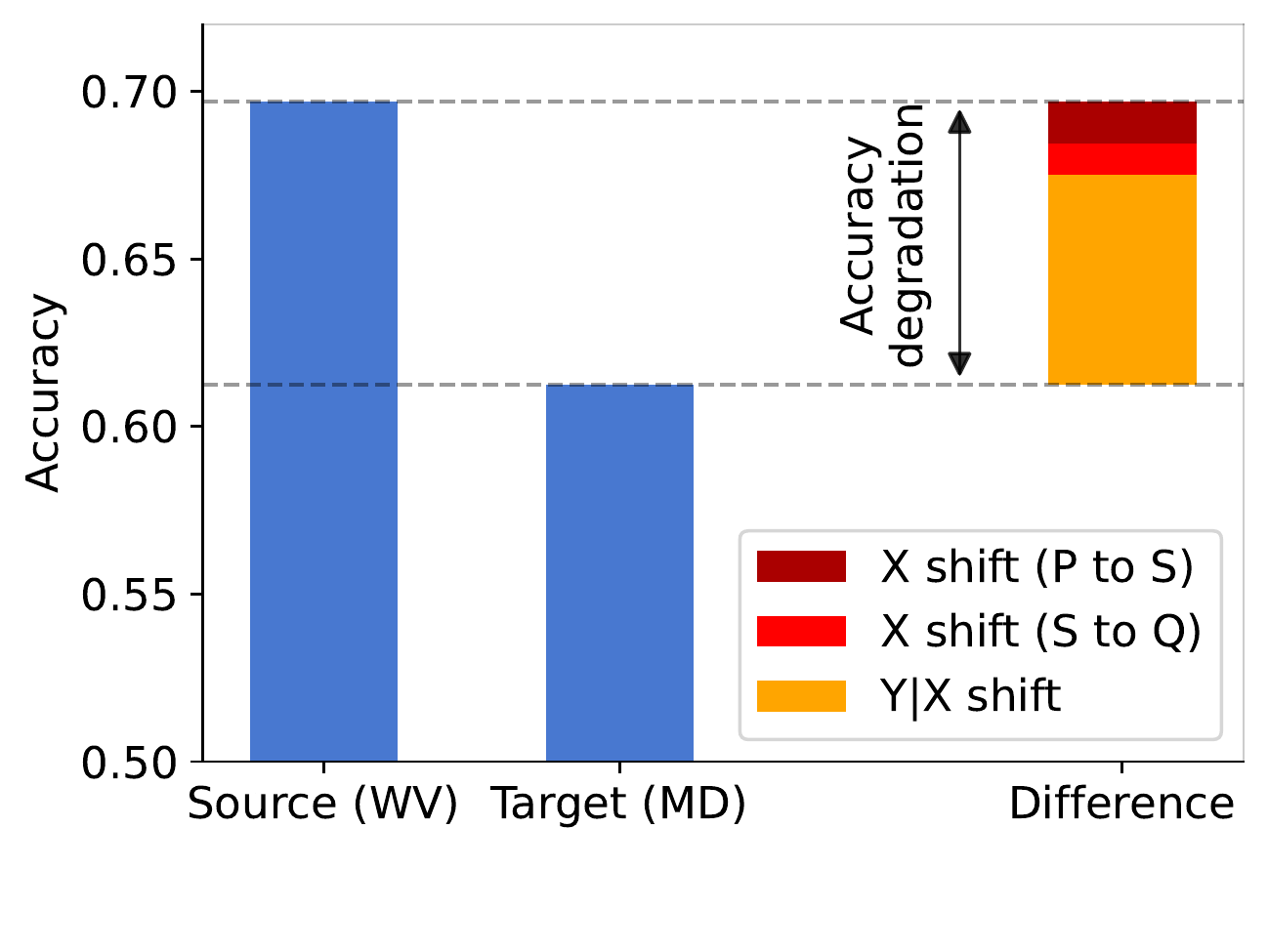}
            \caption{$Y\mid X$ shift: original model trained on West Virginia
            and 
            evaluated on Maryland}
            \label{fig:y_shift_missing}
\end{wrapfigure}
Due to the cost of collecting and labeling enough data, it may be challenging
to train a new model from scratch for each target site.  Consider a prediction
model trained on data collected from one site (West Virginia), which
may be applied to other target sites (Maryland). To construct a specific type
of $Y \mid X$ shift between sites, we consider a scenario where the initial
feature set in the training data did not include educational attainment data.
People in Maryland tend to be more educated than people in West Virginia, and
educational attainment affects employment.  Consequently, when educational
attainment is not included in $X$, a $Y\mid X$ shift arises from marginalizing
out the effect of education over a different distribution in West Virginia
versus Maryland.

Let us now take the perspective of a modeler
who did not know how this distribution shift was constructed. 
In Figure \ref{fig:y_shift_missing}, observe that the prediction model
trained in West Virginia performs worse in Maryland than in West Virginia; our
goal is to
understand this difference in performance. The decomposition depicted in the
figure attributes substantial loss to a $Y\mid X$ shift. In response to such a
decomposition, one might consider reasons why there is a $Y \mid X$ shift,
including the possibility of a shift in the distribution of important
unobserved variables. 
Analysts could
use domain knowledge to consider educational attainment as one such unobserved
variable, and obtain and append this feature to the original training set.
Notably, collecting more features can often be less costly
than collecting enough data from Maryland to fit a new model, and has the additional benefit of improving the model accuracy overall.
Indeed, when we fit a new model on West Virginia incorporating educational
attainment, we find that the new model performs well on both West Virginia and
on Maryland, with an accuracy of 70\% on Maryland, which is an improvement over the
previous 61\%.

\subsubsection{$X$ shift: selection bias in age}
\label{sec:adult_x_oversample}

Data can often be collected with unrecognized selection bias. 
Here, we consider a data
collection process that oversamples people
up to the age of 25 (e.g. from sampling near a college campus).  
We expect that it is easier to predict whether people of
age $\le 25$ are employed, as most of them are still in school.
In contrast, older adults
may encounter new situations that affect employment. 
Let us take the perspective of a modeler who notices
that there is a difference between the training data and the general
population (target $Q$) after the model is trained, and who obtains a 
carefully collected dataset over the general population for evaluation
purposes. We demonstrate our approach on two training distributions that
differ in their magnitude of selection bias and illustrate how the $X$ shift
terms
in the decomposition~\eqref{eq:decomp} can reflect such differences,
and consequently inform future modeling interventions.

\begin{figure}[t]
    \centering
    \hspace{-.9cm}
    \subfloat[\centering Model trained on data with age $\leq $25]{\label{fig:x_shift_new}{\includegraphics[height=5.7cm]{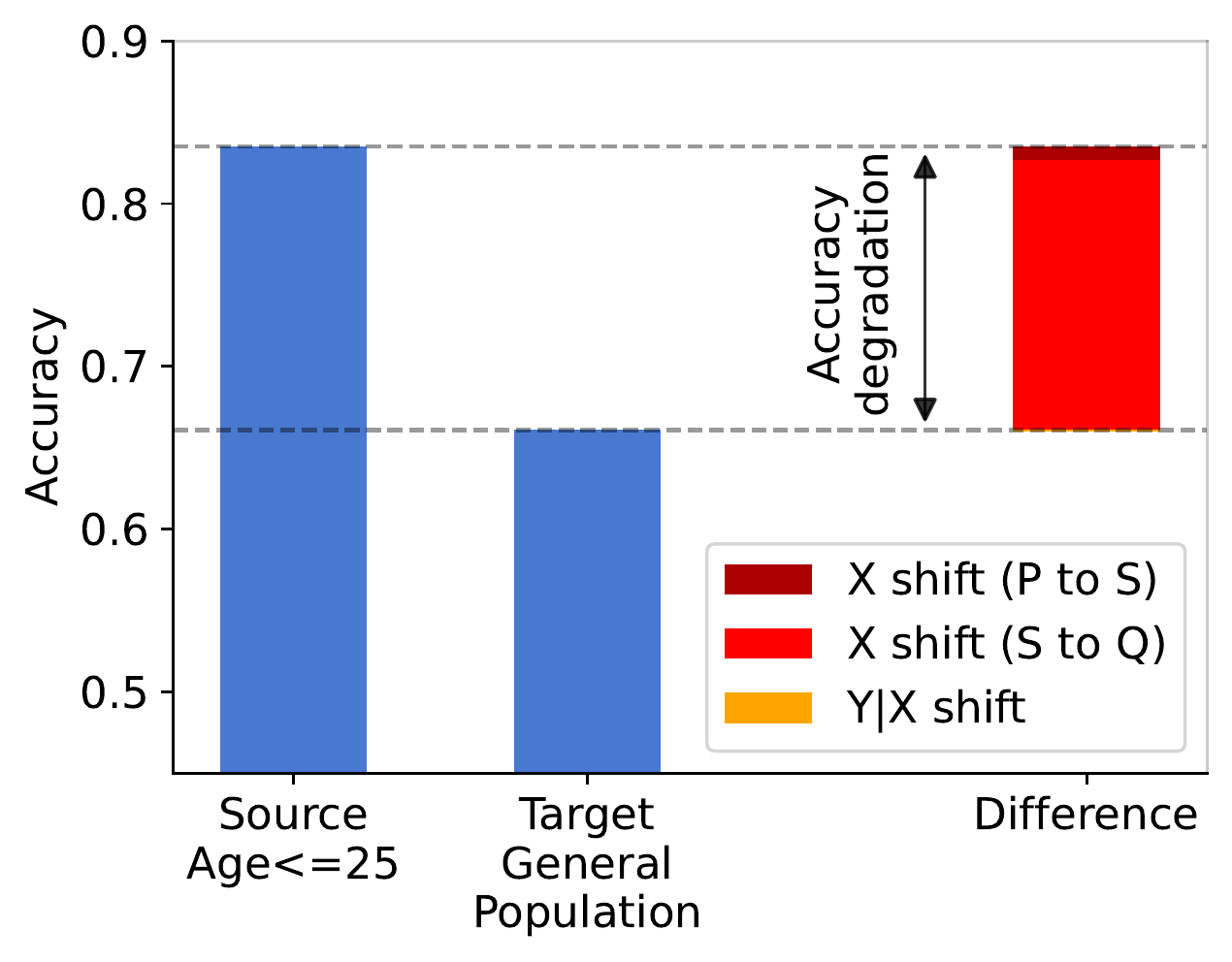}}
    }%
      \hspace{.5cm}
      \subfloat[\centering Model trained on data over-sampling age
      $\leq $25]{\label{fig:x_shift_oversample}{\includegraphics[height=5.7cm]{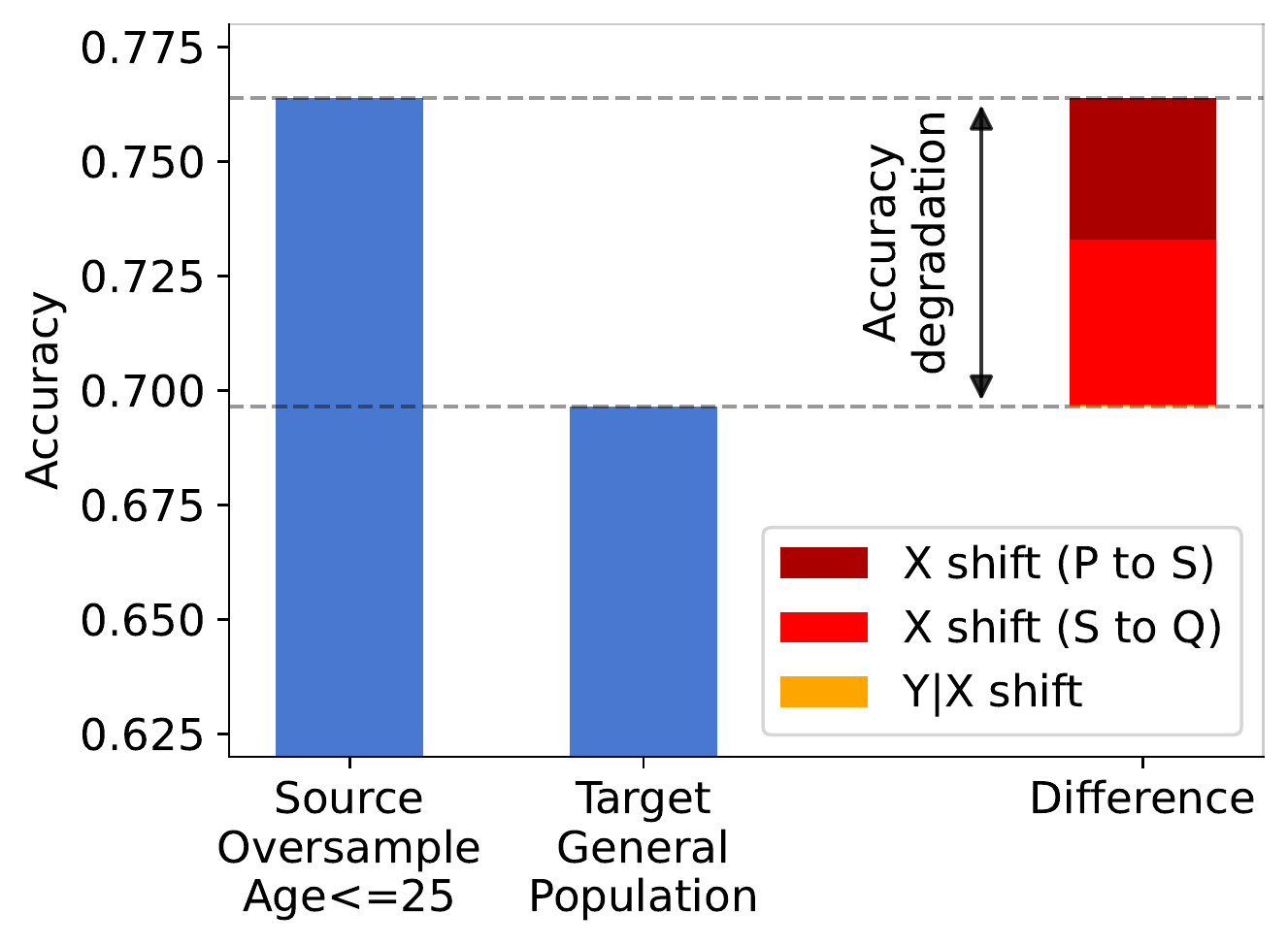}
        }}%
      \vspace{.4cm}
      \caption{\label{fig:x_shift} Moderate vs. severe $X$-shift. Both models are
        evaluated on the general population.}
  \end{figure}

We first focus on the extreme case where the training data \emph{only}
consists of people of up to age 25. In Figure \ref{fig:x_shift_new}, the trained
model performs much worse on the general population, and our decomposition
correctly reveals that the decrease in performance is due to a $X$
shift.  The performance degradation is primarily attributed to $X$ shift ($S\to
Q$),
indicating that the model performed poorly on examples in the general
population ($Q$) that were rarely encountered during training. To better
understand the shift from $P_X$ to $Q_X$, we use the feature importance
method for random forests \cite{PedregosaVaGrMiThGrBlPr11} on the domain classifier
and find that the most
important feature, as expected, is age. Our method thus suggests that data
collection over an older population may be necessary in order to perform
well on the general population. Indeed, a new
employment model fitted on more samples from all ages has an improved
target accuracy of 73\%, compared to the original model's target accuracy of 66\%.

Next, we consider a less dramatic shift in the distribution of $X$, where
people up to the age of $25$ are merely \emph{overrepresented} in the training
data. 
In Figure \ref{fig:x_shift_oversample}, we observe that the model
performs worse on the general population as before, and our decomposition
again reveals that the decrease in performance is primarily attributed to an
$X$ shift.  
Unlike in the previous example, the training data here includes people
of all ages, and we see that the $X$ shift ($P\to S$) term is larger as a result. This suggests
that instead of resorting to an expensive data collection process, reweighting
the original dataset may be an effective algorithmic solution to adapt to the
target. We fit a new
model on an appropriately re-weighted version of the original training data
and find that the new model achieves an accuracy of 80\% on the target, a
significant improvement over the previous model's target accuracy of 69\%.

\begin{figure}[t]
    \centering
    \includegraphics[width=0.9\linewidth]{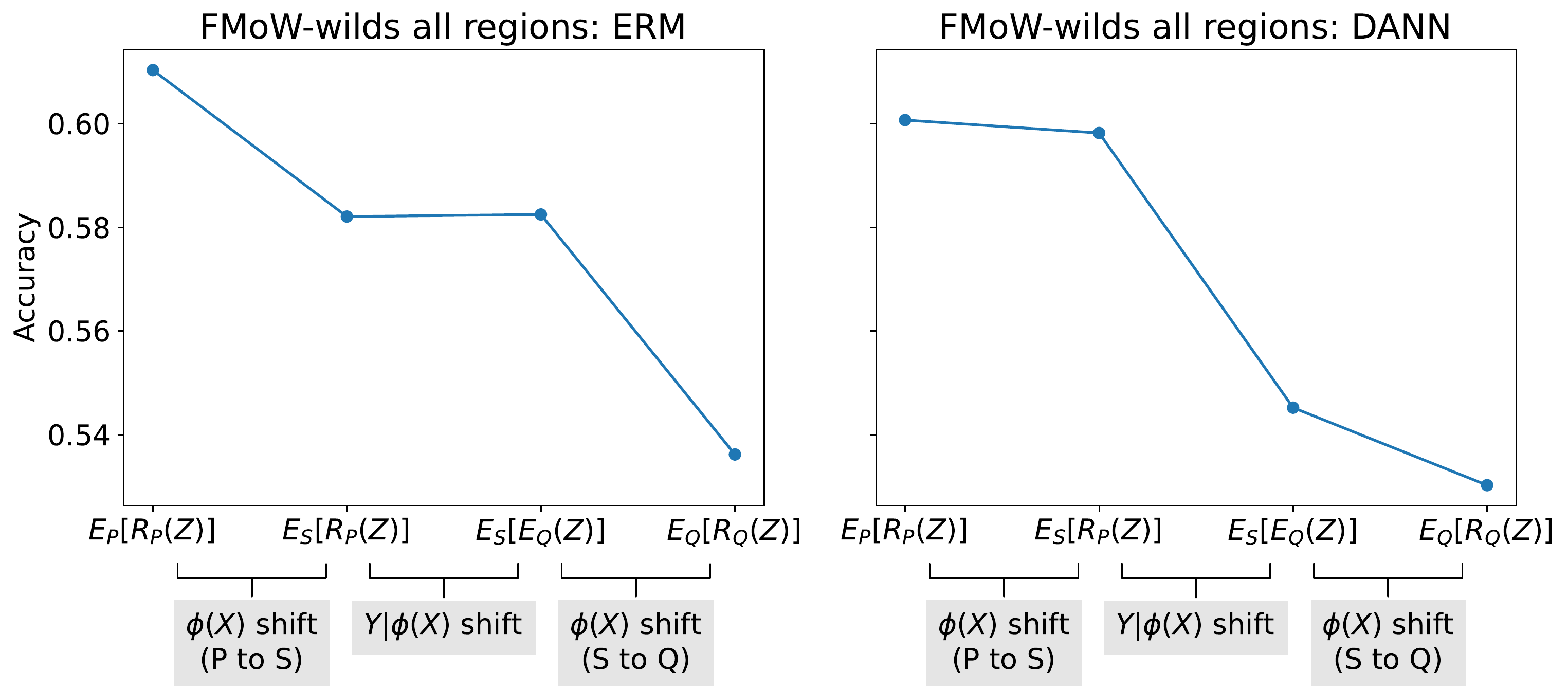}
    \vspace{0.5em}
    \caption{
           ERM (left) vs DANN (right) in the distribution shift from FMoW-wilds. 
        On the vertical axis is model accuracy; all results hold using accuracy in
        place of $\ell(\cdot)$. 
        On the horizontal axis are the four values
        in Equation \eqref{eq:decomp-gen}. Their consecutive differences are 
        the $Z=\phi(X)$ shift ($P\to S$), $Y\mid \phi(X)$ shift, and $\phi(X)$ shift ($S\to Q$)
        terms, respectively. %
        Both ERM and DANN perform worse on target than training, and 
        DANN is not better on target data compared to ERM.
        For ERM,
        the $\phi(X)$ shift terms are largest;
        for DANN, the $Y\mid \phi(X)$ terms are largest. 
    }
    \label{fig:wilds_erm_dann}
\end{figure}

\subsection{Diagnosing failures of domain-adaptation methods}
\label{sec:wilds}

We study an image classification problem where the learned feature
representation plays a critical role in the distributional
robustness of a prediction model. 
We focus on satellite image classification,
which can inform humanitarian and sustainability efforts by tracking
deforestation, population density, and other economic
metrics~\cite{SagawaKoLeGaXiShKuHuYaMa21}. 
In order to track these metrics effectively, 
such models must perform reliably over space and time.
We consider a spatio-temporal distribution shift for
classifying land use from satellite images on a variant of the Functional Map
of the World dataset \cite{SagawaKoLeGaXiShKuHuYaMa21}, called FMoW-wilds
\cite{ChristieFeWiMu18}.  The training data consists of images before 2013,
while the target data is from 2016 and after, and the proportion of samples
from different geographical regions is different between training and target.

\citet{SagawaKoLeGaXiShKuHuYaMa21} recently observed that classifiers trained
via the standard empirical risk minimization (ERM) approach (also known as
sample average approximation) perform significantly worse on the target
distribution compared to training. They also noted that algorithmic
interventions do not, in fact, provide meaningful robustness gains over ERM. To
illustrate how these methods can fail, we focus on a
specific domain adaptation method, called Domain Adversarial Neural Network
(DANN)~\cite{GaninUsAjGeLaLaMaLe16}. The main idea of DANN is to learn a
feature embedding $\phi(X)$ of $X$ such that the distribution of $\phi(X)$ is
difficult to distinguish between training and target. Specifically,
DANN
uses labeled $(X,Y)$ data from training and unlabeled $X$ data from target to learn
neural network feature representations $\phi(X)$ that are not only useful for
predicting $Y$ on the training data,
but are also difficult to classify as being
from training vs target. 
Intuitively, 
requiring $\phi(X)$ to be 
difficult to classify
as being from either distribution can reduce $\phi(X)$
shift, and thus also loss attributed to $\phi(X)$ shift.

We investigate why models trained using DANN perform poorly on the target,
even though DANN
was designed to perform well across a distribution shift. Using the output of the
penultimate layer of the neural network as features so that
$Z_{\rm ERM}=\phi_{\rm ERM}(X)$ and
$Z_{\rm DANN}=\phi_{\rm DANN}(X)$ for the respective models, we 
fit $\hat\pi$ using a logistic regression and present
our decomposition~\eqref{eq:decomp-gen} in
Figure~\ref{fig:wilds_erm_dann}. Compared to the ERM model, the DANN
counterpart has less performance degradation attributed to $\phi(X)$ shift.
However, the performance degradation attributed to $Y\mid \phi(X)$ 
is much larger for the DANN model, so that DANN performs poorly overall,
even though it successfully has less performance degradation attributed to
$\phi(X)$ shift. 
Such a performance degradation can occur if
the assumption that the distributions
of $\phi(X)$ should be indistinguishable between train and target is 
inappropriate. 
This trade-off between different types of distribution shifts
highlights the pitfalls of 
solely focusing on protecting against $\phi(X)$-shifts, as is
common practice in domain adaptation.
See Section~\ref{sec:theory} for confidence intervals.

%% file: theory.tex
\section{Statistical properties}
\label{sec:theory}

In this section, we study the statistical properties of the estimators
$\hat\theta_P,\;\hat\theta_Q$~\eqref{eq:estimate_theta_q}, and the estimators
for the decomposition terms from Equation~\eqref{eq:decomp} from Algorithm
\ref{alg:algo}.  We show asymptotic normality, efficiency, and the validity of
bootstrapping methods for estimating confidence intervals.  Then, we apply our
statistical guarantees to calculate confidence intervals for the experiments
in Section~\ref{sec:experiments}.

We give conditions under which we know that $\hat\theta_P, \;\hat\theta_Q$,
and the estimators for the decomposition terms from Equation~\eqref{eq:decomp}
from Algorithm \ref{alg:algo} are asymptotically normal.  Knowing the
asymptotic distribution allows us construct confidence intervals for the true
parameters $\theta_P,\;\theta_Q$, and the decomposition terms from
Equation~\eqref{eq:decomp}.  In \Appendix{}~\ref{sec:efficiency}, we use
heuristics developed by \citet{Kennedy22} to show that our estimator is
asymptotically efficient, suggesting that our estimator cannot be improved
upon in terms of asymptotic variance.

Recall our notation from Section~\ref{sec:estimation}
\begin{align*}
    \begin{split}
  & \alpha^*~~\mbox{is the proportion of the pooled data that comes from}~ Q_\featuresub \\
  & T = \begin{cases}
    0 ~~&\mbox{if}~\tilde{\feature}~\mbox{is from}~P_\featuresub \\
    1 ~~&\mbox{if}~\tilde{\feature}~\mbox{is from}~Q_\featuresub
    \end{cases}\\
    & \pi^*(\featureval) \defeq \P(T=1\mid\tilde{\feature}=\featureval)
  =\frac{\alpha^* q(\featureval)}{\alpha^* q(\featureval)+(1-\alpha^*) p(\featureval)}
    \end{split}
\end{align*}
The statistical behavior of $\hat\theta_P,\; \hat\theta_Q,$ and the estimators
for the decomposition terms from Equation~\eqref{eq:decomp} depend on the
estimator $\hat\pi(\featureval)$ used for the nuisance parameter
$\pi^*(\featureval)$. However, as is the case for many other semiparametric
estimators, when the nuisance parameter estimates converge sufficiently
quickly and smoothly, the asymptotic distributions of $\hat\theta_P$ and
$\hat\theta_Q$ are well-approximated in a way that does not depend on the form
of the nuisance parameter estimator \cite{Newey94}. In particular, the
nuisance parameters need not be estimated at the $\sqrt{n}$-rate for
$\hat\theta_{P},\; \hat\theta_Q$ to be asymptotically normal at the
$\sqrt{n}$-rate. We show asymptotic normality of the main estimators when the
nuisance parameter $\pi^*(\cdot)$ can only be estimated at the $n^{1/4}$-rate.
Key to this, however, is ``undersmoothing'' of the nuisance parameters \citep{NeweyMc94}, as we describe below.

In this section, we focus on understanding the approximation when the nuisance
parameter is estimated using nonparametric kernel smoothing. Kernel smoothing is
a well-established nonparametric estimation technique \cite{NeweyMc94}, where the
prediction for a test point $\featureval$ is a locally-weighted average of the
outcomes for observations ``near'' $\featureval$ in the training data. The weighting is
defined by a kernel function $K(\cdot)$ and a bandwidth $\sigma > 0$, 
and the weighting is done using $K_\sigma(t):=\sigma^{-1} K(t/\sigma)$. 
Applying this approach to estimate $\pi^*(\cdot)$ gives
\begin{equation*}
  \hat\pi(\featureval) = \frac{\frac{1}{n}\sum_{i=1}^n
    T_iK_\sigma(\featureval-\feature_i)}{\frac{1}{n}\sum_{i=1}^n
    K_\sigma(\featureval-\feature_i)}.
\end{equation*}
While other estimators will give similar results under different assumptions
about the data generating process, focusing on this one allows us to take
advantage of existing technical results for kernel smoothing and to focus our
attention on the behavior of the semiparametric estimators themselves.

We write the kernel smoothing estimate $\hat\pi(\cdot)$ as the ratio of
two kernel density estimators: 
let $A=[1,T]^\top$
and \begin{align}
    \label{eq:hat_gamma}
    \hat \gamma (\featureval)
    &= \frac{1}{n}\sum_{i=1}^n A K_\sigma(\featureval-\feature_i)
\end{align}
so that $\hat\gamma$ is an estimator for the true $\gamma^*$ with
$\gamma_j^*(\featureval):=m_\featuresub(\featureval)\E[A_j\mid
\feature=\featureval]$, with
$\gamma_1^*(\featureval)=m_\featuresub(\featureval)$ the marginal density of
$z$ across the pooled data from $P$ and $Q$, and
$\gamma_2^*(\featureval)=m_\featuresub(\featureval)\E[T\mid
\feature=\featureval]$.  Then
$\hat \pi(\featureval)=\hat\gamma_2(\featureval)/\hat\gamma_1(\featureval)$.

To get good rates of convergence,
we need to be careful with choosing the kernel and bandwidth.
\begin{assumption}[Kernel density estimator assumptions]
    \label{ass:kernel}
Let $d$ be the dimension of $X$. For some choice of constants $k$ and $p$, 
    \begin{enumerate}
        \item $K(u)$ is differentiable of order $p$, the derivatives
            of order $p$ are bounded, $K(u)$ is zero outside of a bounded set,
            $\int K(u)du=1$, there is a positive integer $k$
            such that for all $j<k$, $\int K(u)[\bigotimes_{l=1}^j u]du=0$.
        \item
            The bandwidth $\sigma=\sigma(n)$ satisfies
            $n\sigma^{2d+4p}/(\ln n)^2\to \infty$ and $n\sigma^{2k}\to 0$.
    \end{enumerate}
\end{assumption}
\begin{rmk}
    If the bandwidth $\sigma=\sigma(n)=n^b$, then for Assumption
    \ref{ass:kernel} to hold, we would need
    \begin{align*}
    -\frac{1}{2r+4d}<b<-\frac{1}{2k}.
    \end{align*}
    This requires the bandwidth to shrink faster than the rate used to get minimax optimal convergence of the nuisance parameters themselves with kernel smoothing \citep{Stone82, NeweyMc94}. As a result, the pointwise variance of the nuisance parameter estimates will be higher, and the bias will be lower. This turns out to be critical for getting $\sqrt{n}$-consistency of the semiparametric estimators $\hat{\theta}_{Q}$ and $\hat{\theta}_{P}$, and is one widely applicable lesson that applies broadly to most applicable methods for nuisance parameter estimation in our setting. Therefore, we recommend users to choose hyperparameters that lead to mild, yet notiable undersmoothing when fitting nuisance parameters for our method. Unfortunately, reliable methods for automatically choosing hyperparameters to achieve these requirements are currently not well-known in the literature.
  \end{rmk}

To ensure that the estimates from nonparametric kernel smoothing converge
sufficiently quickly over the entire domain, we additionally make the following assumptions.
\begin{assumption}\
\label{ass:dgp}
\begin{enumerate}
    \item $\feature $ has support on a compact set $\mathcal \feature$, on which its density
        $m_\featuresub(\featureval)$ is bounded above and below (away from 0):
        there are 
        $B_{mL}>0$ and
        $B_{mU}<\infty$ such that
        $B_{mL}< m_\featuresub(\featureval)<B_{mU}$ 
        for all $\featureval$.
    \item $\pi^*(\featureval):=\P(T=1\mid \feature=\featureval)$ is bounded away from 0 and
        1:
	  there is a $\delta_\pi>0$ such that
        $0<\delta_\pi<\pi^*(\featureval)<1-\delta_\pi <1$ for
	  almost every $\featureval$.\footnote{As a consequence, $\alpha:=\P(T=1)$ also satisfies
          $0<\delta_\pi<\alpha<1-\delta_\pi<1$.}
      \item $\pi^*(\featureval)$ is continuous almost everywhere. \tc{is this on
          $\mathcal X$?}
        \item There is a version of $\gamma^*(\featureval)$ that is continuously
            differentiable to order $p$ with bounded derivatives on an open set
            containing $\mathcal X$. \tc{previously, part of the kernel 
            assumptions}
\end{enumerate}
\end{assumption}
\noindent Assumptions like these are standard for kernel smoothing estimators
\citep{NeweyMc94, Tsybakov09}, and are necessary to ensure convergence over
the entire domain.

With these assumptions, the nuisance parameters will not typically be
estimated at the parametric $\sqrt{n}$ rate. However, when plugged into our
semiparametric method, the overall estimators $\hat\theta_P$ and
$\hat\theta_Q$ will converge at a $\sqrt{n}$ rate, and be asymptotically
normal, so long as the loss $L=\ell(f(X),Y)$ is sufficiently regular. We
describe sufficient regularity conditions for $L$ in the assumption below.
\begin{assumption}\
\label{ass:loss-reg}
\begin{enumerate}
    \item $\E[L^4]< B_{L^4}$ for some $B_{L^4} < \infty$. %
    \item $\mu_Q(\featureval):=\E[L\mid T=1,\feature=\featureval]$ and $\mu_P(\featureval):=\E[L\mid T=0,\feature=\featureval]$ are continuous
    almost everywhere.
    \item $\mu_Q(\featureval), \mu_P(\featureval) < B_\mu <\infty$ for some $B_\mu$ on $\mathcal X$. 
\end{enumerate}
\end{assumption}

Because our estimators~\eqref{eq:estimate_theta_q} depend on the product of
functions of $\hat{\pi}(\cdot)$ and the loss $L$,
Assumption~\ref{ass:loss-reg} ensures that small estimation errors in
$\hat{\pi}(\cdot)$ aren't amplified in pathological ways when multiplied by
$L$.  With these three assumptions, our estimators are jointly regular and
asymptotically linear. Our main result (Theorem~\ref{thm:if_estimator} to
come) makes precise the asymptotic distribution of the estimator
$\hat\theta_Q$. A symmetric result holds for $\hat\theta_P$, and from these we
can deduce the asymptotic distribution of the decomposition terms in
Equation~\eqref{eq:decomp}.

To state Theorem~\ref{thm:if_estimator}, we introduce additional
notation. Building on Equation \eqref{eq:estimation_m}, let $M(\cdot)$ denote
the distribution of the observed data.  Let $M_\featuresub(\cdot)$ be the
marginal distribution of $\feature$ under this distribution, and
$m_\featuresub(\featureval)$ its density. As before, define $T$ as a binary
dummy variable that is $1$ when the observation comes from the target
distribution $Q$ and $0$ when it comes from the training distribution $P$.
Then as in Section~\ref{sec:estimation} we can write
\begin{align*}
    \theta_Q %
    = \E_{M}\left[ \frac{\dif{S_\featuresub}}{\dif{M_\featuresub}}
    \frac{\indic{T = 1}}{\P_M(T = 1 \mid \feature)} L \right]
    ~~\mbox{ and }~~
    \theta_P %
    = \E_{M}\left[ \frac{\dif{S_\featuresub}}{\dif{M_\featuresub}}
    \frac{\indic{T = 0}}{\P_M(T = 0 \mid \feature)} L \right].
\end{align*}
Recalling that $\pi^*(\featureval):=\P_M(T=1\mid
\feature=\featureval)$,\footnote{In this section we write $\pi^*, \gamma^*,
\alpha^*$ to denote true values of nuisance parameters.}
and using Equation \eqref{eq:s} and some calculations, 
\begin{gather}
  \frac{\dif\S_{\featuresub}}{\dif M_{\featuresub}}\propto
    \lambda(\pi^*(\feature), \alpha^*)
  ~~\text{ where}~~
  \label{eq:lambda}
  \lambda(\pi,\alpha) =
  \frac{\pi(1-\pi)}{(1-\alpha)\pi+\alpha(1-\pi)}.
\end{gather}
Then we can write our estimands $\theta_P$, $\theta_Q$ as functionals of the data
distribution $M$, the conditional probability $\pi^*(\feature)$, and the marginal
probability $\alpha^*=\P_M(T=1)$ as
\begin{equation}
  \label{eqn:theta-restated}
    \theta_P =
    \frac{\E_M\left[L\frac{1-T}{1-\pi^*(\feature)}\lambda(\pi^*(\feature),\alpha^*)
    \right]}{\E_M\left[\frac{1-T}{1-\pi^*(\feature)}\lambda(\pi^*(\feature),\alpha^*)\right]}
    ~~~\mbox{and}~~~
    \theta_Q = \frac{\E_M\left[L
    \frac{T}{\pi^*(\feature)}\lambda(\pi^*(\feature),\alpha^*)
    \right]}{\E_M\left[\frac{T}{\pi^*(\feature)}\lambda(\pi^*(\feature),\alpha^*)\right]}.
\end{equation}
For brevity, let $\Z$ denote $(\feature, T, L)$. Then
we show the asymptotic linearity of $\hat\theta_Q$:
\begin{theorem}
    \label{thm:if_estimator}
    Let $\hat\gamma$ be as in Equation \eqref{eq:hat_gamma}.
    Then, under Assumptions \ref{ass:kernel}, \ref{ass:dgp}, and \ref{ass:loss-reg},
    \begin{align*}
      \sqrt{n}(\hat\theta_Q-\theta_Q)
      =\sum_{i=1}^n \psi_Q(\z_i)/\sqrt{n}+o_P(1),
    \end{align*}
    so that
    \begin{align*}
      \sqrt{n}(\hat\theta_Q-\theta_Q)\cd N(0,\var(\psi_Q(\Z))).
    \end{align*}
    where
    \begin{align*}
      \psi_Q(\z)
      = \left(\E\left[\frac{T}{\pi^*(\feature)}\lambda(\pi^*(\feature),\alpha^*)
        \right]\right)^{-1}
        & \Bigg\{ (l-\theta_Q)\frac{t}{\pi^*(\featureval)}
        \lambda(\pi^*(\featureval),\alpha^*) \\
      & \hspace{1em}
        +
        \E\left[
        (L-\theta_Q)
        \frac{T
        }{\pi^*(\feature)}
        \frac{\partial}{\partial \alpha}
        \lambda(\pi^*(\feature),\alpha)\Big|_{\alpha=\alpha^*}\right]
        (t-\alpha^*) \\
      & \hspace{1em}
        +(\mu_Q(\featureval)-\theta_Q)\pi^*(\featureval)(t-\pi^*(\featureval))
        \frac{\partial}{\partial
        \pi}\left[\frac{\lambda(\pi,\alpha^*)}{\pi}\right]\Big|_{\pi=\pi^*(\featureval)}
        \Bigg\}
    \end{align*}
  \end{theorem}

See \Appendix~\ref{sec:proofs} for the proof. An analogous result holds for $\hat\theta_P$, 
and the decomposition terms from Equation~\eqref{eq:decomp}
follows immediately as $\hat\theta_P,\;
\hat\theta_Q$ are asymptotically linear.
Theorem \ref{thm:if_estimator} can be further generalized: %
\citet{Yadlowsky22}
shows that the assumptions needed to guarantee asymptotic linearity with the
same influence function extend immediately to cross-fitted analogues of these
estimators (see \Appendix{}~\ref{sec:splits} for more on
cross-fitting). Additionally, the literature has many results showing
asymptotic linearity of semiparametric methods using other nuisance parameter
estimators under more general conditions than those assumed in the proofs
provided here \cite{Newey97, ChernozhukovChDeDuHaNeRo18}. This is because many
of these results have a similar form, with the first order effect of estimating
the
nuisance parameters is the influence function in
Theorem~\ref{thm:if_estimator}, and the remaining error terms that depend
on
the specific choice of nuisance parameter estimator are lower order. See, for
example, \citet{Newey94, Newey97, Kennedy22, Yadlowsky22}.

The asymptotic linearity in our result directly ensures that subsampling
procedures such as half-sampling \citep{ChungRo13} are justified for
constructing calibrated confidence intervals. The half-sample bootstrap
\citep{Efron82, PrastgaardWe93} is a simple procedure of this form, which
\citet[Lemma 4]{YadlowskyFlShBrWa22} showed directly applies here
(Corollary~\ref{cor:half-sample}). The half-sample bootstrap is constructed by
repeatedly drawing samples of half the data without replacement, constructing
the estimator $\hat\theta_{Q}^\ast$ with each sample, and analyzing the
distribution of the errors $\hat\theta_{Q}^\ast - \hat\theta_{Q}$. Our
next result shows that this provides be a good approximation of the sampling
distribution of $\hat{\theta}_{Q} - \theta_{Q}$.  In this corollary (of
\citet[Lemma 4]{YadlowskyFlShBrWa22}), $P^\ast$ refers to the observed Monte
Carlo distribution of the procedure, and is conditional on the observed data
(and therefore, $\hat\theta_{Q}$).
\begin{corollary}
\label{cor:half-sample}
Under the conditions of Theorem~\ref{thm:if_estimator}, conditionally on the
observed data, the Monte Carlo distribution of $\sqrt{n}(\hat\theta_{Q}^\ast -
\hat\theta_{Q}) \overset{P^\ast}{\to} N(0, \var(\psi_Q(\Z)))$.
\end{corollary}
\noindent \hn{Link hyperref to proof}
\tc{I think there isn't one; the result follows immediately.}

\paragraph{Confidence intervals in practice}

Showing that the nonparametric bootstrap \citep{Efron82} gives valid
confidence intervals requires additional work. However, given the regularity
of our functional, we conjecture that the nonparametric bootstrap also
provides calibrated confidence intervals. The advantage of the nonparametric
bootstrap is its simplicity and reduced sensitivity to finite sample defects
resulting from each subsample being larger in size. In this subsection, we
empirically verify that the nonparametric bootstrap provides similar inference
to the theoretically-justified half-sample bootstrap, so we recommend this
procedure for routine practice.

\input{adult_ci}

In Table~\ref{tab:adult_sd}, we first compare different ways of calculating
confidence intervals for Section~\ref{sec:adult} using methods from
Section~\ref{sec:theory}.
Next, we calculate nonparametric bootstrap confidence intervals for
Section~\ref{sec:wilds}. The confidence intervals in Table~\ref{tab:wilds} show
that the conclusions drawn from our previous analysis are not due to
happenstance and are statistically sound.

\input{wilds_ci}

%% file: adult_ci.tex
\begin{table}[htp]
    \centering
    \begin{subfigure}[t]{0.80\textwidth}
            \centering
        \input{tables/adult_yx_3.tex}

            \caption{$Y|X$ shift: original model trained on West Virginia
            and 
            evaluated on Maryland}
            \label{tab:y_shift_missing}
    \end{subfigure}

    \begin{subfigure}[t]{0.80\textwidth}
    \centering
    \vspace{1em}
        \input{tables/adult_x_kids_3.tex}

        \caption{$X$ shift: model trained on only age $\leq $25 and
            evaluated on general population}
            \label{tab:x_shift_new}
    \end{subfigure}

    \begin{subfigure}[t]{0.8\textwidth}
            \centering
    \vspace{1em}
        \input{tables/adult_x_oversample_3.tex}

\caption{$X$ shift: model trained on over-sampling age $\leq $25 and
            evaluated on general population}
            \label{tab:x_shift_oversample}
    \end{subfigure}
    \vspace{1em}
    \caption{
        Point estimate and estimates of standard error
        for estimators of decomposition terms
        for Section \ref{sec:adult}.
        The standard error column names are abbreviated: 
        ``NP boot'' is short for nonparametric boostrap and ``HS boot'' is short
        for half-sample bootstrap, as described in Section~\ref{sec:theory}.
    Each type of bootstrapping is done with 500 bootstrap re-samples.
        ``IF est'' is short for plug-in estimator of influence function:
        we calculate the standard error using influence functions as in
        results like Theorem~\ref{thm:if_estimator}
        and plug in estimates $\hat\pi(x),\hat\mu_Q(x),\hat\alpha$ in
        place of $\pi^*(x),\mu_Q(x),\alpha^*$.  
    }
\label{tab:adult_sd}
\end{table}

%% file: tables/adult_yx_3.tex
        \begin{tabular}{lcccc}
            \toprule
            & & \multicolumn{3}{c}{Standard error} \\
            \cmidrule(lr){3-5}
            Term & Estimate & NP boot & HS boot & IF est \\
            \midrule
            $X$ shift ($P\to S$) & 0.78\%  & 0.66\%  & 0.71\%  & 0.96\% \\
 $Y\mid X$ shift & 6.29\%  & 0.97\%  & 1.17\%  & 1.13\% \\
            $X$ shift ($S\to Q$) & 1.35\%  & 0.68\%  & 0.75\%  & 0.86\% \\        
            \bottomrule
        \end{tabular}

%% file: tables/adult_x_kids_3.tex
        \begin{tabular}{lcccc}
            \toprule
            & & \multicolumn{3}{c}{Standard error} \\
            \cmidrule(lr){3-5}
            Term & Estimate & NP boot & HS boot & IF est \\
            \midrule
            $X$ shift ($P\to S$) & 0.23\%  & 0.38\%  & 0.43\%  & 0.53\% \\
 $Y\mid X$ shift & -0.01\%  & 0.43\%  & 0.53\%  & 0.56\% \\
            $X$ shift ($S\to Q$) & 17.21\%  & 0.33\%  & 0.34\%  & 0.40\% \\        
            \bottomrule
        \end{tabular}

%% file: tables/adult_x_oversample_3.tex
        \begin{tabular}{lcccc}
            \toprule
            & & \multicolumn{3}{c}{Standard error} \\
            \cmidrule(lr){3-5}
            Term & Estimate & NP boot & HS boot & IF est \\
            \midrule
            $X$ shift ($P\to S$) & 2.03\%  & 0.79\%  & 0.67\%  & 1.13\% \\
 $Y\mid X$ shift & 0.27\%  & 0.97\%  & 1.05\%  & 1.26\% \\
            $X$ shift ($S\to Q$) & 5.15\%  & 0.64\%  & 0.63\%  & 0.89\% \\        
            \bottomrule
        \end{tabular}

%% file: wilds_ci.tex
\begin{table}[!htbp]
    \centering
    \input{tables/wilds_estimates_and_boot.tex}

    \vspace{0.5em}
    \caption{Estimates for ERM vs DANN, corresponding to Figure
    \ref{fig:wilds_erm_dann}. The standard errors are calculated using non-parametric
    bootstrap with 100 bootstrap samples (``NP boot SE'')}
    \label{tab:wilds}
\end{table}

%% file: tables/wilds_estimates_and_boot.tex
        \begin{tabular}{lccccc}
            \toprule
            & \multicolumn{2}{c}{ERM} & & \multicolumn{2}{c}{DANN}\\
            \cmidrule(lr){2-3} \cmidrule(lr){5-6}
            Term &  Estimate & NP boot SE & & Estimate & NP Boot SE\\
            \midrule
            $\phi(X)$ shift ($P\to S$)  & 2.82\%  & 0.15\%  &  & 0.25\%  & 0.13\% \\
            $Y\mid \phi(X)$ shift & -0.04\%  & 0.40\%  &  & 5.29\%  & 0.40\% \\
            $\phi(X)$ shift ($S\to Q$) & 4.62\%  & 0.15\%  &  & 1.49\%  & 0.13\% \\        
            \bottomrule
        \end{tabular}

%% file: discussion.tex
\section{Discussion}
\label{section:discussion}
Distribution shift is a fundamental problem in data-driven decision-making.
Still, there are limited tools for understanding changes in model performance
with respect to different types of distribution shifts.
We introduce a simple method, DIstribution Shift DEcomposition (DISDE) 
that can diagnose out-of-distribution
performance by attributing the change in loss to three terms, corresponding to
1) an increase in harder but frequently
  seen examples from training, 2) changes in the relationship between
  features and outcomes, and 3) poor performance on examples infrequent or
  unseen during training (Sections \ref{sec:decomp}, \ref{sec:estimation}). 
We empirically demonstrate how 
DISDE can inform modeling improvements on tabular and image data (Section \ref{sec:experiments}), 
and we
analyze its asymptotic properties (Section \ref{sec:theory}). In this final section, we situate the current work in the large
body of literature on distribution shifts (Section~\ref{section:related}) and
discuss the limitations of our approach alongside future research directions
(Section~\ref{section:future}).

\subsection{Related work}
\label{section:related}

Distribution shift is an important topic across multiple research
communities. Much of the work is on training better models through
methodological improvements.  Like ours, some works focus on taxonomies for
discussing distribution shifts \cite{LiptonWaSmo18, ScholkopfJaPeSgZhMo12,
  TranLiDuPhCoRe22}.  Complementary to our work, other works focus on
collecting data across distribution shifts.  However, there is not much
research on attributing changes in model performance to different types of
distribution shifts.  Concurrent to our work, \citet{ZhangSiGhJo22} use
Shapley value to attribute changes in performance to various types of shift,
but do not address differences in support or density between distributions as
we do with the shared distribution in DISDE (Section
\ref{sec:decomp}). 

In the machine learning community, domain adaptation methods train models using
data from a training distribution, and aim to perform well on a specified target distribution.  Standard
approaches assume the $Y \mid X$ distribution is fixed and reweight training data
to resemble the target distribution~\cite{Shimodaira00, HuangGrBoKaScSm07,
  BickelBrSc07, SugiyamaKrMu07, TsuboiKaHiBiSu09}.  When we expect shifts in
the $Y\mid X$ distribution,~\citet{MeinshausenBu15, RothenhauslerBuMePe18} propose
learning models that have good performance against causal interventions that
affect $Y \mid X$.  More recently, several authors have studied approaches that
aim to learn feature representations $\phi(X)$ such that $Y \mid \phi(X)$,
or functionals thereof, is
invariant across multiple environments~\cite{PetersBuMe16, SaenkoKuFrDa10,
ArjovskyBoGuLo19, RosenfeldRaRi20}.

There is a large body of work on distributionally robust optimization (DRO)
methods that optimize performance over distributions that are ``close'' to the
training distribution. While the majority of these works consider shifts in
the joint distribution $(X, Y)$~\cite{EldarBeNe04, DelageYe10, XuCaMa12,
  Ben-TalHeWaMeRe13, LamZh15, MiyatoMaKoNaIs15, BertsimasGuKa18, Lam19,
  StaibJe19, BertsimasGuKa18, VanParysEsKu21, DuchiNa21}, some notions of
closeness (e.g., Wasserstein distance) can flexibly handle both joint and
covariate shifts~\cite{Shafieezadeh-AbadehEsKu15, BlanchetKaZhMu17, GaoChKl17,
  BlanchetKaMu19, KuhnEsNgSh19}. Despite the extensive literature on DRO,
limited work study particular types of distribution shifts. Focusing on the
case of covariate shifts,~\citet{DuchiHaNa22} develop tailored methods that
optimize worst-case subpopulation shifts over a set of
covariates. Similarly,~\citet{SahooLeWa22} propose a related worst-case
subpopulation formulation over $Y\mid X$-shifts, holding the marginal distribution
of $X$ fixed.

The causal inference community takes a more nuanced approach to distribution
shift.  There is extensive work on sensitivity analysis
(e.g.,~\cite{Rosenbaum10, Rosenbaum11, KallusZh18, Fogarty19, KallusMaZh19,
  YadlowskyNaBaDuTi22}), which studies the amount of unobserved
confounding---analogous to the magnitude of $Y \mid X$ shift in predictive
scenarios---to call into question the conclusion of observational studies.
When a known target population differs from the study population, a large body
of research adjusts observations to resemble the target~\cite{StuartCoBrLe11,
  Tipton13, KernStHiGr16, LeskoBuWeEdHuCo17}. These approaches are analogous
to domain adaptation under covariate shift; analogous to distributionally
robust optimization,~\citet{JeongNa22} recently formulate worst-case
sensitivity approaches for covariate shift. More generally, the external
  validity of a study can be called into question due to different
distribution shifts.~\citet{EgamiHa22} formalize and contextualize the type of
shifts that naturally arise in social sciences.

Our approach complements design-based and benchmarking research in statistics
and machine learning.  Collecting maximally heterogeneous data is a
classical idea in experimental design, with a recent focus on multi-site
designs~\cite{CrucesGa07, BanerjeeKaZi15, GertlerShAlCaMaPa15, DehejiaPoSa21,
  TiptonPe17, TiptonRo18}. In machine learning, building on older 
literature on distribution shifts~\cite{Hand06, Quinonero-CandelaSuScLa08},
there is renewed interest in benchmarking model performance on unseen
test distributions different from training~\cite{RechtRoScSh19, KohSaEtAl20,
  HendrycksBaMuKaWaDoDeZhPaGu21, MillerKrReSc20}. For example,
~\citet{RechtRoScSh19, TaoriDaShCaReSc20} observe a trend
in which models trained on the same data universally suffer a relative
performance degradation on new distributions, regardless of the training
method and model capacity.  By evaluating model robustness over multiple
distributions, these works spur empirical progress by simulating
typical ML deployment processes where models encounter a priori unknown
distributions under operation. While out-of-distribution benchmarks and industrial monitoring
systems primarily focus on detecting performance degradation~\cite{Huyen22},
we build a diagnostic that generates qualitative insights on the cause of the
observed failure. We hope this work provides a starting point for building a
principled language for understanding performance degradation over distribution
shifts.

\subsection{Limitations and future work}
\label{section:future}

Our work has several limitations, some of which are by design.  For example,
by design, DISDE does not directly propose new interventions; it is a
diagnostic that can help decide between potential interventions, or to help
assess existing interventions.  In addition, DISDE only applies in certain
settings: for now, it only concerns %
shifts between pairs of distributions, and not among more distributions.  It
also only applies to losses that can be written as
$\E[\ell(f(X),Y)]$. Therefore, it does not directly apply to an F1-score or a
worst-subgroup loss, for example.  Lastly, our work assumes the existence of a
shared distribution $S_\featuresub$ (Section \ref{sec:decomp}).  If there is
no shared support between $P_\featuresub$ and $Q_\featuresub$, then our
decomposition will not be valid.

Other limitations of our approach concern the limitations of the work that it
builds on.  In particular, our methodology involves an auxiliary domain
classifier, which is analogous to the propensity score in causal
inference. How best to learn a propensity score is a field of active research.
For example, there are questions on the role of balancing covariates between
treatment and control, versus modeling treatment assignment
\cite{ImaiRa14,Zhao2019,ChattopadhyayHaZu2020}. 
In our work we use a very simple way
to learn a propensity score and leave extensions to future work.

Broadly speaking, our work highlights the potential benefits of an operations
engineering approach to the training, deployment, and use of machine learning
models. Classical quality control and reliability engineering methods have had
major impacts across multiple industries (e.g., car
manufacturing~\cite{PyzdekKe03}, electronics manufacturing~\cite{Miner45},
software development processes~\cite{Paulk93}).  As the use of ML models
becomes increasingly prevalent and complex, we will need ways to diagnose
model failures and to direct resources to the appropriate modeling
interventions.  However, currently, even the most advanced industrial
monitoring systems can only detect performance degradation, without
attribution to its cause~\cite{Huyen22}.  We hope this paper spurs further
work to build rigorous and scalable tools for the quality management of ML
applications. For example, to analyze the local sensitivity of a simulation
output, several authors have used conditional variances to attribute variance
due to different subsets of the
input~\cite{HommaSa96,SaltelliRaAnCaCaGaSaTa08,Owen14,SongNeSt16} and related
approaches may yield fruit in the context of ML models.

%% file: ack.tex
\section{Acknowledgements}
Tiffany Cai is supported by the National Science Foundation Graduate Research
Fellowship under Grant No. DGE-2036197. Hongseok Namkoong was partially
supported by the Amazon Research Award.

%% file: theory_proofs.tex
\section{Proofs for asymptotic properties for the estimator}
\label{sec:proofs}

In this section, we will prove Theorem \ref{thm:if_estimator}, to show the
asymptotically linear expansion of $\hat\theta_Q$.  Recalling the definition
of $\theta_Q$~\eqref{eqn:theta-restated}, denote by $D_Q$ and $N_Q$ the
denominator and the numerator 
\begin{align*}
    D_Q =\E\left[ \frac{T}{\pi\opt(\feature)}\lambda(\pi\opt(\feature), \alpha\opt) \right] 
    ~~~\text{and}~~~
    N_Q =    \E\left[L \frac{T}{\pi\opt(\feature)}\lambda(\pi\opt(\feature),
\alpha\opt) \right].
\end{align*}
Observe that the numerator and denominator of $\theta_Q$ (and also
$\hat\theta_Q$) are nearly identical: we can think of the denominator $D_Q$
(and its empirical plug-in $\what{D}_Q$) as a special case of the numerator
$N_Q$ (and also $\hat N_Q$) but with $ L \defeq 1$. Below, we focus on
asymptotic expansions for $N_Q$ without loss of generality.

The following lemma shows it suffices to find the asymptotically linear
expansion of the numerator and denominator of $\hat\theta_Q$ separately. We
defer its proof to Section~\ref{sec:proof-of-lem-fractions}.
\begin{lemma}
\label{lem:fractions}
Assume that $A_n\cp A$, $B_n\cp B$, $\sqrt{n}(A_n - A) = \frac{1}{\sqrt{n}} \sum_{i=1}^n a_i + o_P(1)$ and $\sqrt{n}(B_n - B) = \frac{1}{\sqrt{n}} \sum_{i=1}^n b_i + o_P(1)$. Additionally, assume that $B > 0$. Then,
\begin{equation*}
\sqrt{n}\left(\frac{A_n}{B_n} - \frac{A}{B}\right) = \frac{1}{\sqrt{n}}
    \sum_{i=1}^n \left(\frac{1}{B}a_i - \frac{A}{B^2}b_i\right) + o_P(1).
\end{equation*}
\end{lemma}
\noindent Because of Lemma \ref{lem:fractions}, it suffices to prove the
following proposition; we give its proof in the rest of this section.
Let $\what{N}_Q$ be the empirical plug-in for $N_Q$
\begin{align}
  \label{eq:hat-N_Q}
  \what{N}_Q := 
  \sum_{i=1}^n l_i \frac{t_i}{\hat\pi(\featureval_i)}\lambda(\hat\pi(\featureval_i),\hat\alpha).
\end{align}
\begin{proposition}
  \label{prop:if_numerator}
  $\sqrt{n}(\hat{N}_Q - N_Q) = \frac{1}{\sqrt{n}}\sum_{i=1}^n \psi_{N_Q}(\z_i)
  + o_{P}(1)$, where
  $\psi_{N_Q}(\z) = g(\z) + h(\z) + \delta(\z)$ and
  \begin{align*}
    g(\z)
    &= l\frac{t}{\pi^*(\featureval)}\lambda(\pi^*(\featureval),\alpha^*)-N_Q \\
    h(\z)
    &=\E\left[
      L\frac{T}{\pi^*(\feature)}\frac{\partial}{\partial\alpha}
      \lambda(\pi^*(\feature),\alpha) \Big|_{\alpha=\alpha^*}
      \right] 
      (t-\alpha^*) \\
    \delta(\z)
    &= \mu_Q(\featureval) \pi^*(\featureval) 
      (t-\pi^*(\featureval))
      \frac{\partial}{\partial
      \pi}\left[\frac{\lambda(\pi,\alpha^*)}{\pi}\right]\Big|_{\pi=\pi^*(\featureval)}.
  \end{align*}
\end{proposition}

To begin, recall that $\pi(\featureval)=\gamma_2(\featureval)/\gamma_1(\featureval)$ and define
\begin{align}
\label{eq:g}
    g(\z,\gamma,\alpha) &= l\frac{t}{\pi(\featureval)}\lambda(\pi(\featureval),\alpha) -N_Q
\end{align}
so that
\begin{align*}
    \what{N}_Q-N_Q=n^{-1} \sum_{i=1}^n g(\z_i,\hat\gamma,\hat\alpha)
.\end{align*}
Since $g$ is continuously differentiable with respect to $\alpha$,
expand 
$\hat\alpha$ around $\alpha^*$ as 
\begin{align*}
    g(\z_i, \hat\gamma,\hat\alpha)
    &=
    \nabla_\alpha g(\z_i,\hat\gamma,\overline
    \alpha)(\hat\alpha-\alpha^*)
    +
    g(\z_i,\hat\gamma,\alpha^*)
\end{align*}
where $\overline\alpha$ is a mean value. 
By Lemma~\ref{lem:G_alpha} below,
$n^{-1}\sum_{i=1}^n 
\nabla_\alpha
g(\z,\hat\gamma,\alpha)
\cp
    \E[
        \nabla_\alpha
        g(\z,\gamma^*,\alpha)
    ]
.$
\begin{lemma}
\label{lem:G_alpha}
Under the assumptions and definitions of Theorem
    \ref{thm:if_estimator},
\begin{align*}
    n^{-1}\sum_{i=1}^n \nabla_\alpha
g(\z,\hat\gamma,\overline\alpha)\cp
    \E[
        \nabla_\alpha g(\z,\gamma^*,\alpha^*)]=:G_\alpha
\end{align*}
\end{lemma}
\noindent Proof of this lemma is deferred to Section~\ref{sec:proof-of-lem-G_alpha}.
Noting that $\hat\alpha=n^{-1}\sum_{i=1}^n t_i$ and 
letting 
\begin{align*}
  h(\z) :=\E[\nabla_\alpha g(\z,\gamma^*,\alpha^*)]
  (t-\alpha^*)
  = \E\left[
  L\frac{T}{\pi^*(\feature)}\nabla_\alpha \lambda(\pi^*(\feature),\alpha^*)
  \right] 
  (t-\alpha^*),
\end{align*}
we arrive at
\begin{align}
\label{eq:if_alpha}
    \sum_{i=1}^n 
    g(\z_i, \hat\gamma,\hat\alpha)
    /\sqrt{n}&=
    \sum_{i=1}^n 
    [h(\z_i)
    +g(\z_i,\hat\gamma,\alpha^*)
    ]/\sqrt{n}
    +o_P(1)
.\end{align}

To find an asymptotically linear representation of $\sum_{i=1}^n g(\z_i,\hat\gamma,\alpha^*)/\sqrt{n},$
we will show that under the assumptions made in Theorem~\ref{thm:if_estimator},
$\sum_{i=1}^n g(\z_i,\hat\gamma,\alpha^*)/\sqrt{n}$ satisfies the assumptions of \citet[Theorem 8.11]{NeweyMc94}, which we restate now for convenience. Note that while the theorem as stated in \citet{NeweyMc94} only shows asymptotic normality, in the proof they show asymptotic linearity, so we have modified the result to state the asymptotic linearity result, instead.
We have also modified the result for the case where $g$ is linear in the
estimated parameter, in
which case consistency of the estimated parameter does not need to be shown
\cite{NeweyMc94}. 

\begin{assumption}[{\citet[Assumptions 8.1-8.3 and assumption from Theorem
    8.11]{NeweyMc94}}]
\label{ass:8.1-8.3}
Let $d$ be the dimension of $X$. 
    \begin{enumerate}
        \item $K(u)$ is differentiable of order $p$, the derivatives
            of order $p$ are bounded, $K(u)$ is zero outside of a bounded set,
            $\int K(u)du=1$, there is a positive integer $m$ 
            such that for all $j<m$, $\int K(u)[\bigotimes_{l=1}^j u]du=0$.
        \item There is a version of $\gamma^*(\featureval)$ that is continuously
            differentiable to order $p$ with bounded derivatives on an open set
            containing $\mathcal \feature$.
        \item There is $r\geq 4$ such that $\E[\|A\|^r]<\infty$
            and $\E[\|A\|^r|\feature=\featureval]m_\featuresub(\featureval)$ is bounded. %
        \item
            The bandwidth $\sigma=\sigma(n)$ satisfies
            $n\sigma^{2d+4p}/(\ln n)^2\to \infty$ and $n\sigma^{2m}\to 0$.
    \end{enumerate}
\end{assumption}

\begin{assumption}[{\citet[assumptions from Theorem 8.11]{NeweyMc94}}]
    \label{ass:8.11_unnamed}
    Let $\beta$ be the parameter of interest, and $\hat \beta$ its estimator
    where
    $\hat \beta-\beta=\frac{1}{n}\sum_{i=1}^n g(\Z_i,\hat \gamma)$. 
    \begin{enumerate}
        \item $\E[g(\Z,\gamma^*)]=0$
        \item $\E[\|g(\Z,\gamma^*)\|^2]<\infty$
        \item $\mathcal \feature$ is a compact set.
    \end{enumerate}
\end{assumption}
\begin{assumption}[{\citet[enumerated assumptions from Theorem
    8.11]{NeweyMc94}}]
    \label{ass:8.11}
    There is a vector of functionals $G(\z,\gamma)$ that is linear in $\gamma$
    such that
    \begin{enumerate}[(i)]
        \item
    for $\|\gamma-\gamma^*\|$ small
            where the norm is the Sobolev norm\footnote{
    $\| \gamma \|:= \max_{\ell\leq p} \sup_{\featureval\in\mathcal \feature} \| \partial^\ell
            \gamma(\featureval)/\partial^\ell \featureval \|$},
    $\|g(\z,\gamma)-g(\z,\gamma^*)-G(\z,\gamma-\gamma^*)\|\leq
    b(\z)\|\gamma-\gamma^*\|^2$, and
    $\E[b(\Z)]<\infty$;
\item $\|G(\z,\gamma)\|\leq c(\z)\|\gamma\|$ and
    $\E[c(\Z)^2]<\infty$;
\item there is $v(\featureval)$ with $\int G(\z,\gamma)dM(\z)=\int
    v(\featureval)\gamma(\featureval)d\featureval$
    for all $\|\gamma \|<\infty$;
\item $v(\featureval)$ is continuous almost everywhere,
    $\int \|v(\featureval)\| d\featureval<\infty$, and there is $\epsilon>0$
    such that
    $\E[\sup_{\|\nu \|\leq \epsilon}\|v(\feature+\nu )\|^4]<\infty$.
    \end{enumerate}
    \end{assumption}
\begin{theorem}[N+M Theorem 8.11]%
    \label{thm:8.11}
    Let $\gamma^*$ be the nuisance parameter and $\hat
    \gamma$ its kernel density estimate satisfying
    Assumption \ref{ass:8.1-8.3}.
    Let $\beta$ %
be the parameter of interest, and $\hat \beta$ its estimator
    satisfying Assumption \ref{ass:8.11_unnamed}. 
    Assume there is a vector of functionals $G(\z,\gamma)$
    satisfying Assumption \ref{ass:8.11}.
    Then for $\delta(\z)=v(\featureval)a-\E[v(\featureval)a]$,
    \begin{align*}
\sum_{i=1}^n g(\z_i,\hat\gamma)/\sqrt{n}
=
        \sum_{i=1}^n [g(\z_i,\gamma^*)+\delta(\z_i)]/\sqrt{n}+o_P(1).
    \end{align*}
 \end{theorem}

We now proceed to use this result to prove the asymptotically linear representation 
\begin{align}
\label{eq:if_gamma}
    \sum_{i=1}^n 
    g(\z_i,\hat\gamma,\alpha^*) 
    /\sqrt{n}
    &=
    \sum_{i=1}^n
    [g(\z_i,\gamma^*,\alpha^*) + \delta(\z_i)]/\sqrt{n}
    +o_P(1)
\end{align}
for our choice of $g(\z, \gamma, \alpha)$.
Then, the desired result follows from combining Equations \eqref{eq:if_gamma}
and \eqref{eq:if_alpha},
and then applying Lemma \ref{lem:fractions}.
What remains is to check the conditions of Theorem \ref{thm:8.11}. 
\\
\newline
    \noindent \textbf{Verifying Assumption \ref{ass:8.1-8.3}:}
    \begin{enumerate}
        \item Assumed in Assumption \ref{ass:kernel}
        \item Assumed in Assumption \ref{ass:dgp}
        \item There is $r\geq 4$ such that $\E[\|A\|^r]<\infty$
            and $\E[\|A\|^r|\feature=\featureval]m_\featuresub(\featureval)$ is bounded: %
            recall that $A=[1,T]$
            and $T$ takes values in $\{0,1\}$.
            Then let $r=4$, $\|A\|^r\leq 2$, so that 
            $\E[\|A\|^r]\leq 2<\infty$, and 
            and $\E[\|A\|^r\mid \feature=\featureval]m_\featuresub(\featureval) \leq 2
            B_{mU}<\infty$ 
            by Assumption \ref{ass:dgp}.
        \item Assumed in Assumption \ref{ass:kernel}
    \end{enumerate}

\noindent\textbf{Verifying Assumption \ref{ass:8.11_unnamed}:}
\begin{enumerate}
    \item $\E[g(\Z,\gamma^*,\alpha^*)]=0$: this holds by definition of $g$
        (Equation \eqref{eq:g}) and
        $N_Q$ (Equation \eqref{eq:hat-N_Q}).
    \item $\E[|g(\Z,\gamma^*,\alpha^*)|^2]<\infty$:
        \begin{align*}
            \E[|g(\Z,\gamma^*,\alpha^*)|^2]&=
\E\left[
    \left|L\frac{T}{\pi^*(\feature)}\lambda(\pi^*(\feature),\alpha^*) -N_Q\right|^2
\right]       
.\end{align*}
        This is finite since $N_Q<\infty$, and then by Cauchy-Schwarz since 
        $\E[L^4]<B_{L^4}$ by
        Assumption \ref{ass:loss-reg},
and 
        $\left|\frac{T}{\pi^*(X)}\lambda(\pi^*(X),\alpha^*)\right|^4$ is
        bounded, since
        $T\in\{0,1\}$ and
        $$\frac{\lambda(\pi^*(X),\alpha^*)}{\pi^*(X)}=\frac{1-\pi^*(\feature) }{(1-\alpha^*)\pi^*(\feature)+\alpha^*(1-\pi^*(\feature)) }\leq
        \frac{1-\delta_\pi}{\delta_\pi}$$ with $\delta_\pi$ from Assumption 
        \ref{ass:dgp}.

    \item Assumed in Assumption \ref{ass:dgp}.
\end{enumerate}

\noindent    \textbf{Verifying Assumption \ref{ass:8.11}:}
    \begin{enumerate}[(i)]
        \item
    For $\|\gamma-\gamma^*\|$ small\footnote{the norm on $\|\gamma-\gamma^*\|$ is the Sobolev norm},
    $\left|g(\z,\gamma,\alpha^*)-g(\z,\gamma^*,\alpha^*)-G(\z,\gamma-\gamma^*)\right|\leq
    b(\z)\|\gamma-\gamma^*\|^2$, and
    $\E[b(\Z)]<\infty$:
            By Taylor-expanding $g$ around $\gamma^*(\featureval)$ (where in the
            following
            equations
            we abuse notation and also use $g$ to mean
            $g(w,\gamma(\featureval),\alpha)$, where the
            second argument is the value of the function $\gamma$ evaluated at $\featureval$,
            rather than the function $\gamma$.
            We also write
            $\nabla_\gamma g(\cdot)$ to denote the
            derivative of this new $g$ with respect to its second argument, the
            value $\gamma(\featureval)$, rather than to
            the function $\gamma(\cdot)$, and similarly for $\nabla^2_\gamma
            g(\cdot)$),
            \begin{align*}
                g(\z,\gamma,\alpha^*)&=g(\z,\gamma^*(\featureval), \alpha^*)+\nabla_\gamma
                g(\z,\gamma^*(\featureval), \alpha^*)^\top (\gamma(\featureval)-\gamma^*(\featureval))
                \\&\hspace{1em}
                +\frac{1}{2}
                (\gamma(\featureval)-\gamma^*(\featureval))^\top \nabla_\gamma^2
                g(\z,\overline m, \alpha^*)(\gamma(\featureval)-\gamma^*(\featureval))
            \end{align*}
            where $\overline m$ is a mean value on the line between
            $\gamma(\featureval)$ and $\gamma^*(\featureval)$. 
            Note that $g$ is twice differentiable with respect to $\gamma(\featureval)$ in an open set containing this
            line (since $\|\gamma-\gamma^*\|$ small, so that $\pi$ is bounded away from
            0 and 1) so that the expansion holds.
            Thus 
            let $G(\z,\gamma-\gamma^*)$ be the first-order term in the expansion
            above: %
            \begin{align*}
                G(\z,\gamma)&:=
                \nabla_\gamma g(\z,\gamma(\featureval), \alpha^*)^\top \gamma(\featureval)
                \\&=lt
                \frac{\partial}{\partial\pi}\left(
                    \frac{\lambda(\pi,\alpha^*)}{\pi}\right
                )\Big|_{\pi=\pi^*(\featureval)} \nabla_\gamma \pi(\gamma^*(\featureval))^\top  \gamma(\featureval)
                \\&=
                lt
                \frac{\partial}{\partial\pi}\left(
                    \frac{\lambda(\pi,\alpha^*)}{\pi}\right
                )\Big|_{\pi=\pi^*(\featureval)} m_\featuresub(\featureval)^{-1} [-\pi^*(\featureval),1]  \gamma(\featureval)
            \end{align*}
            where we used
            $\pi(\gamma(\featureval))=\gamma_2(\featureval)/\gamma_1(\featureval)$,
            $\gamma_1(\featureval)=m_\featuresub(\featureval)$
            is the marginal density of $\feature$,
            and 
            \begin{align*}
                \frac{\partial}{\partial\pi}\left(
                    \frac{\lambda(\pi,\alpha^*)}{\pi}\right
                )\Big|_{\pi=\pi^*(\featureval)}
                =
-\frac{1-\alpha^*}{((1-\alpha^*)\pi^*(\featureval)+\alpha^*(1-\pi^*(\featureval)))^2}
            .
            \end{align*}%
            Then to verify the assumption, 
            \begin{align*}
                &\left|g(\z,\gamma,\alpha^*)-g(\z,\gamma^*,\alpha^*)-G(\z,\gamma-\gamma^*)\right|
                \\&=\frac{1}{2}
                \left|
                (\gamma(\featureval)-\gamma^*(\featureval))^\top \nabla_\gamma^2
                g(\z,\overline m, \alpha^*)(\gamma(\featureval)-\gamma^*(\featureval))
                \right|
                \\&\leq \frac{1}{2} 
                \left\|
                 \nabla_\gamma^2
                g(\z,\overline m, \alpha^*)
                \right\|_F \|\gamma-\gamma^*\|^2
            .\end{align*}
            Thus to verify the assumption, let $b(\z) = \frac{1}{2}\left\|\nabla_\gamma^2
            g(\z,\overline
            m, \alpha^*)\right\|_F$. Some simple calculus shows that 
                 $\nabla_\gamma^2
                g(\z,\overline m, \alpha^*)
                =lt \phi(\overline m)
            $
            where
            \begin{align*}
                \phi(\overline m) &=
                \frac{\partial^2}{\partial\pi^2}\left(
                    \frac{\lambda(\pi,\alpha^*)}{\pi}\right
                )\Big|_{\pi=\overline \pi}
                \nabla_\gamma \pi(\overline m) \nabla_\gamma \pi(\overline m)^\top
                +
                \frac{\partial}{\partial\pi}\left(
                    \frac{\lambda(\pi,\alpha^*)}{\pi}\right
                )\Big|_{\pi=\overline \pi}
                \nabla_\gamma^2 \pi(\overline m)
            .\end{align*}%
            Then $\E[\|\phi(\overline m)\|_F^2]$ is bounded: 
            since
            $\|\gamma-\gamma^*\|$ is small, 
            $\overline m$ is close to $\gamma^*(\featureval)$,
            so that $\overline m_1$ is bounded away from 0, and $\overline\pi$
            is bounded away from both 0 and 1, 
            so that each term in $\phi(\overline m)$ is bounded, so that $\E[\|\phi(\overline m)\|_F^2]$
            is bounded. Then, applying Cauchy-Schwarz,
            $\E[b(\Z)]\le\frac{1}{2}\sqrt{\E[ (LT)^2 ]\E[\|\phi(\overline m)\|_F^2
            ]}<\infty$ as $\E[(LT)^2]\leq \E[L^2]\leq \E[L^4]<B_{L^4}$ by Assumption \ref{ass:loss-reg}.

        \item $|G(\z,\gamma)|\leq c(\z)\|\gamma\|$ and
            $\E[c(\z)^2]<\infty$:
\begin{align*}
|G(\z,\gamma)|&=
   lt\left|
                \frac{\partial}{\partial\pi}\left(
                    \frac{\lambda(\pi,\alpha^*)}{\pi}\right
                )\Big|_{\pi=\pi^*(\featureval)}
m_\featuresub(\featureval)^{-1}[-\pi^*(\featureval),1]
    \gamma(\featureval)\right|
    \\&\leq 
    lt
    \left|
                \frac{\partial}{\partial\pi}\left(
                    \frac{\lambda(\pi,\alpha^*)}{\pi}\right
                )\Big|_{\pi=\pi^*(\featureval)}
    \right|
    m_\featuresub(\featureval)^{-1}
    \|[-\pi^*(\featureval),1]\| \|\gamma(\featureval)\|
    \\&\leq 
    lt
    \left|
                \frac{\partial}{\partial\pi}\left(
                    \frac{\lambda(\pi,\alpha^*)}{\pi}\right
                )\Big|_{\pi=\pi^*(\featureval)}
    \right|
    m_\featuresub(\featureval)^{-1}
    \sup_{\featureval\in\mathcal \feature}\left\{\|[-\pi^*(\featureval),1]\|\right\} \|\gamma\|
    \\&\leq 
   lt C \|\gamma\|
\end{align*}
for some constant $C$ 
since $
    \left|
                \frac{\partial}{\partial\pi}\left(
                    \frac{\lambda(\pi,\alpha^*)}{\pi}\right
                )\Big|_{\pi=\pi^*(\featureval)}
    \right|
    $ is bounded as in the previous part, 
    and $m_\featuresub(\featureval)^{-1}, \pi^*(\featureval)$ are bounded by
    Assumption \ref{ass:dgp}.
    Here, the norm $\|\gamma\|$ is the Sobolev norm
    while $\|\gamma(\featureval)\|$ is the Euclidean norm.  
    Thus let $c(\z)=ltC$, and $\E[c(\Z)^2]=C^2\E[L^2T^2]\leq C^2\sqrt{\E[(LT)^4]}\leq
    C^2\sqrt{\E[L^4]}\leq C^2\sqrt{B_{L^4}}<\infty$ 
    by Assumption \ref{ass:loss-reg}.

        \item
            There is $v(\featureval)$ with $\int G(\z,\gamma)dM(\z)=\int
            v(\featureval)\gamma(\featureval)d\featureval$:

            Define $v(\featureval)$ by rewriting $\int G(\z,\gamma) dM(\z)$:
\begin{align*}
\int G(\z,\gamma) dM(\z) 
&=\int  lt \frac{\partial}{\partial \pi}\left[\frac{\lambda(\pi,\alpha^*)}{\pi}\right]
\Big|_{\pi=\pi^*(\featureval)}
    m_\featuresub(\featureval)^{-1}[-\pi^*(\featureval),1]\gamma(\featureval) dM(\z)
    \\&= \int \mu_Q(\featureval) \pi^*(\featureval)\frac{\partial}{\partial \pi}\left[\frac{\lambda(\pi,\alpha^*)}{\pi}\right]\Big|_{\pi=\pi^*(\featureval)}
[-\pi^*(\featureval),1]\gamma(\featureval)d\featureval
\end{align*}
where the last equality is essentially obtained by using iterated expectations
to rewrite
$$\E[ LT \xi(\feature)] = \E[ \E[L\mid T=1,\feature]\E[T\mid \feature]\xi(\feature)]$$
since $T\in\{0,1\}$, for $\xi(\featureval) = \frac{\partial}{\partial
\pi}\left[\frac{\lambda(\pi,\alpha^*)}{\pi}\right]\Big|_{\pi=\pi^*(\featureval)}
[-\pi^*(\featureval),1]\gamma(\featureval)$
so that
\begin{align*}
    v(\featureval) = \mu_Q(\featureval)\pi^*(\featureval) \frac{\partial}{\partial \pi}\left[\frac{\lambda(\pi,\alpha^*)}{\pi}\right]\Big|_{\pi=\pi^*(\featureval)}[-\pi^*(\featureval),1].
\end{align*}
        \item
    $v(\featureval)$ is continuous almost everywhere,
    $\int \|v(\featureval)\| d\featureval<\infty$, and there is $\epsilon>0$
    such that
    $\E[\sup_{\|\nu \|\leq \epsilon}\|v(\feature+\nu )\|^4]<\infty$:

            $v(\featureval)$ is continuous almost everywhere since it is the product of
            functions that are continuous almost everywhere: $\pi^*(\featureval)$ and
            $\mu_Q(\featureval)$ are continuous almost everywhere by Assumptions
            \ref{ass:dgp} and \ref{ass:loss-reg},
and $\frac{\partial}{\partial \pi}\left[\frac{\lambda(\pi,\alpha^*)}{\pi}\right]\Big|_{\pi=\pi^*(\featureval)}$
is also continuous in $\featureval$.
            
            $\int \|v(\featureval)\|d\featureval<\infty$: 
            $\|v(\featureval)\|$ is bounded on $\mathcal \feature$ since it is the product of
            several terms that are each bounded on $\mathcal \feature$: 
            $\mu_Q(\featureval)$ is bounded by Assumption \ref{ass:loss-reg},
            $\pi^*(\featureval)$ is bounded by Assumption \ref{ass:dgp}, and
$\frac{\partial}{\partial \pi}\left[\frac{\lambda(\pi,\alpha^*)}{\pi}\right]\Big|_{\pi=\pi^*(\featureval)}$
            is bounded since $\delta_\pi < \pi^*(\featureval) < 1-\delta_\pi$
            for $\delta_\pi>0$ from in Assumption
            \ref{ass:dgp}. 
            The integral is finite since $\mathcal \feature$ is compact. 

           The sup condition is satisfied since $\|v(\featureval)\|$ is
           bounded in $\mathcal \feature$.

\end{enumerate}
Now that we have satisfied the conditions of 
Theorem~\ref{thm:8.11} and we have $v(\featureval)$, let $a=[1,t]^\top$ and define
            \begin{align*}
                \delta(\z)
                &=v(\featureval)a-\E[v(\feature)A]
                \\&=
                \mu_Q(\featureval)
                (t-\pi^*(\featureval))
                \pi^*(\featureval) \frac{\partial}{\partial \pi}\left[\frac{\lambda(\pi,\alpha^*)}{\pi}\right]\Big|_{\pi=\pi^*(\featureval)}
.            \end{align*}
Then by Theorem~\ref{thm:8.11}, 
\begin{align*}
    \sum_{i=1}^n 
    g(\z_i,\hat\gamma,\alpha^*) 
    /\sqrt{n}
    &=
    \sum_{i=1}^n
    [g(\z_i,\gamma^*,\alpha^*) + \delta(\z_i)]/\sqrt{n}
    +o_P(1)
\end{align*}
as desired.

\section{More proofs for asymptotic properties of the estimator}

\subsection{Proof of Lemma~\ref{lem:fractions}}
\label{sec:proof-of-lem-fractions}
Define $r(a,b) = a/b$. 
It is continuously differentiable, so
we can apply the mean value theorem to its derivative, so that for some choices of values
$\overline A, \overline B$ between $A_n$ and $A$, and $B_n$ and $B$,
respectively,
\begin{align*}
    \sqrt{n}\left(\frac{A_n}{B_n} - \frac{A}{B}\right) &=
    \sqrt{n}(r(A_n, B_n)-r(A, B))
\\&=\sqrt{n}\nabla r(\overline{A}, \overline{B})^\top \begin{bmatrix} A_n-A \\ B_n - B\end{bmatrix}
\\&=[1/\overline B, -\overline A/\overline B^2]
\begin{bmatrix}
    \sum_{i=1}^n a_i/\sqrt{n}+o_P(1)\\
    \sum_{i=1}^n b_i/\sqrt{n}+o_P(1)
\end{bmatrix}
.\end{align*}
Since $A_n \cp A$ and $B_n \cp B$, we have that $\overline{A} \cp A$
and $\overline B \cp B$, so
\begin{align*}
\sqrt{n}\left(\frac{A_n}{B_n} - \frac{A}{B}\right)
=\frac{1}{\sqrt{n}}\sum_{i=1}^n 
\left(\frac{1}{B}a_i - \frac{A}{B^{2}}b_i \right)
+o_P(1)
.\end{align*}

\subsection{Proof of Lemma~\ref{lem:G_alpha}}
\label{sec:proof-of-lem-G_alpha}
We will use the following from \citet{NeweyMc94}:
\begin{lemma}[Direct consequence of {\citet[Lemma 8.10]{NeweyMc94}}]
    \label{lemma:8.10}
    If Assumption \ref{ass:8.1-8.3} is satisfied,
    then $\sqrt{n}\|\hat \gamma-\gamma^*\|^2\cp 0$.
\end{lemma}
\noindent Since we are assuming Assumption \ref{ass:8.1-8.3} for showing
Theorem \ref{thm:if_estimator}, the conclusion holds. 

Now, since $\frac{1}{n}\sum_{i=1}^n \nabla_\alpha g(\z,\gamma^*,\alpha^*)\cp
\E[\nabla_\alpha g(\z,\gamma^*,\alpha^*)]$
by law of large numbers, 
    it suffices to show 
    \begin{align*}
        \frac{1}{n}\sum_{i=1}^n\left[\nabla_\alpha g(\z_i,\hat \gamma,\overline
        \alpha)-\nabla_\alpha g(\z_i,\gamma^*,\alpha^*)\right]\cp 0
        .
    \end{align*}
Using the Markov inequality, it suffices to show
    that for $\|\hat\gamma-\gamma^*\|$ small enough, 
    \begin{align*}
           \left| \nabla_\alpha g(\z,\hat \gamma,\overline
        \alpha)-\nabla_\alpha g(\z,\gamma^*,\alpha^*)
        \right|
        \leq b(\z) \left[\|\hat\gamma-\gamma^*\|+\|\hat\alpha-\alpha^*\|\right]
    \end{align*}
    for some $b(\z)$ such that $\E[b(\Z)]<\infty$, since
    $\|\hat\gamma-\gamma^*\|\cp 0$ and $\|\hat\alpha-\alpha^*\|\cp 0$.
Similar to verifying Assumption \ref{ass:8.11}
(where again we 
abuse notation to write $g$ on the RHS to take $\gamma(x)$ 
as the second argument, rather than $\gamma(\cdot)$, and 
we write $\nabla_\gamma$ to denote derivatives with respect to $\gamma(\featureval)$
instead of $\gamma(\cdot)$),  \tc{this notation below: ??}
\begin{align*}
            \nabla_\alpha g(\z,\hat \gamma,\overline
        \alpha)-\nabla_\alpha g(\z,\gamma^*,\alpha^*)
=
    \nabla_{\gamma,\alpha}
            \nabla_\alpha g(\z,\overline m,\overline{\overline \alpha})^\top
            \begin{bmatrix} \hat\gamma(\featureval)-\gamma^*(\featureval) \\ \overline{
                \alpha}-\alpha^*
            \end{bmatrix}
\end{align*}
where $(\overline m, \overline{\overline \alpha})$ is a mean value on the line
between
$(\hat\gamma(\featureval), \overline \alpha)$ and 
$(\gamma^*(\featureval), \alpha^*)$.
Then, note that $\hat\gamma(\featureval)-\gamma^*(\featureval)\leq \|\hat\gamma-\gamma^*\|$.
Let 
$$b(w):=\sup_{\overline m, \overline{\overline \alpha}}\|\nabla_{\gamma,\alpha} \nabla_\alpha g(\z,\overline m,\overline{\overline
\alpha})\|_\infty,$$
so that 
$$
           \left| \nabla_\alpha g(\z,\hat \gamma,\overline
        \alpha)-\nabla_\alpha g(\z,\gamma^*,\alpha^*)
        \right|
        \leq b(\z) \left[\|\hat\gamma-\gamma^*\|+\|\hat\alpha-\alpha^*\|\right]
.$$
Finally, 
$\E[b(W)]<\infty$
by an argument similar to the verification of Assumption \ref{ass:8.11}.

\section{Proofs for efficiency}
\label{sec:efficiency}
We verify that our estimator in Theorem \ref{thm:if_estimator} 
attains the nonparametric efficiency bound, i.e. that
our estimator has the best possible asymptotic variance. 
Instead of rigorously deriving the nonparametric efficiency bound, 
for brevity, we use heuristics from
\cite{Kennedy22}
to obtain the efficient influence function of the estimand.
To show the desired result, we then show the efficient influence 
function of the estimand
is the same as the influence function of the estimator in Theorem
\ref{thm:if_estimator}. 
\subsection{Efficient influence function}
We calculate the efficient influence function for the estimand
using heuristics from \cite{Kennedy22}. 
To simplify notation, we
first deal with only the numerator of $\theta_Q$,
which we denoted as
$$N_Q=\E_M\left[\ell(f(X),Y)\frac{T}{\pi^*(\feature)}\lambda(\pi^*(\feature),\alpha^*)\right]=\E_M\left[\mu_Q(\feature)\lambda(\pi^*(\feature),\alpha^*)\right].$$
To calculate the efficient influence function, 
as in \cite{Kennedy22}, we treat $\mathcal \feature$ as discrete, and use derivative rules 
with simple influence functions as building blocks.
For notational simplicity, we omit $^*$'s for $\pi^*,\alpha^*$ for the
calculation of efficient influence functions.
\begin{align*}
    \IF\{\alpha\}&=T-\alpha
    \\
    \IF\{p(\featureval)\}&=\indic{\feature=\featureval}-p(\featureval)
    \\
    \IF\{p(t)\}&=\indic{T=t}-p(t)
    \\
    \IF\{\pi(\featureval)\}&=\frac{\indic{\feature=\featureval}}{p(\featureval)}(T-\pi(\featureval))
    \\
    \IF\{\mu_Q(\featureval)\} &= \frac{\indic{\feature=\featureval,T=1}}{\P(\feature=\featureval,T=1)}(L-\mu_Q(\featureval))
    =\frac{T\indic{\feature=\featureval}}{p(\featureval)\pi(\featureval)}(L-\mu_Q(\featureval))
    \\
    \IF\{\lambda(\pi(\featureval),\alpha)\}&=\nabla_\pi \lambda(\pi(\featureval),\alpha)\IF\{\pi(\featureval)\}
    +\nabla_\alpha \lambda(\pi(\featureval),\alpha) \IF\{\alpha\}
    \\&=\nabla_\pi \lambda(\pi(\featureval),\alpha)\frac{\indic{\feature=\featureval}}{p(\featureval)}(T-\pi(\featureval))
    +\nabla_\alpha \lambda(\pi(\featureval),\alpha) (T-\alpha)
\end{align*}
Then
\begin{align*}
    \IF\{N_Q\} &= \sum_{\featureval\in\mathcal \feature} \mu_Q(\featureval)\lambda(\pi(\featureval),\alpha)p(\featureval)
    \\&=\sum_{\featureval\in\mathcal \feature} 
\left(\IF\{\mu_Q(\featureval)\} \lambda(\pi(\featureval),\alpha)p(\featureval)+\mu_Q(\featureval)\IF\{\lambda(\pi(\featureval),\alpha)\} p(\featureval)
+\mu_Q(\featureval)\lambda(\pi(\featureval),\alpha)\IF\{p(\featureval)\}
\right)
\\&=
    \sum_{\featureval\in\mathcal \feature} 
\frac{T\indic{\feature=\featureval}}{p(\featureval)\pi(\featureval)}(L-\mu_Q(\featureval)) 
\lambda(\pi(\featureval),\alpha) 
    p(\featureval)
\\&\hspace{1em}
+
    \sum_{\featureval\in\mathcal \feature} 
\mu_Q(\featureval) \pt{\pi} \lambda(\pi,\alpha)%
    \frac{\indic{\feature=\featureval}}{p(\featureval)}(T-\pi(\featureval))p(\featureval)
\\&\hspace{1em}
+
    \sum_{\featureval\in\mathcal \feature} 
    \mu_Q(\featureval)
\pt{\alpha}\lambda(\pi(\featureval),\alpha) (T-\alpha) 
    p(\featureval)
\\&\hspace{1em}
+
    \sum_{\featureval\in\mathcal \feature} 
\mu_Q(\featureval)\lambda(\pi(\featureval),\alpha)(\indic{\feature=\featureval}-p(\featureval))
\\&=
(L-\mu_Q(\feature))
\frac{T}{\pi(\feature)}
\lambda(\pi(\feature),\alpha)
+
\mu_Q(\feature)(T-\pi(\feature))
\pt{\pi} \lambda(\pi,\alpha)%
\\&\hspace{1em}
+
(T-\alpha)
\E\left[ 
\mu_Q(\featureval)
\pt{\alpha}
\lambda(\pi(\featureval),\alpha) 
\right]
+
\mu_Q(\feature)\lambda(\pi(\feature),\alpha)
-N_Q
.
\end{align*}

\subsection{Comparison}
Now we show the efficient influence function from the previous section
is the same as the influence function of the estimator in Theorem
\ref{thm:if_estimator}. We do so by first comparing
the efficient influence function of the numerator of $\theta_Q$, $\IF(N_Q)$,
with the influence function of the estimator for the numerator of $\theta_Q$,
$\psi_{N1}(\z)$.
Then we do the same for the denominator, then $\theta_Q$, then $\theta_P$, then
the terms in the decomposition~\eqref{eq:decomp}. 
We start with the numerator of $\theta_Q$, $N_Q$. 
Recall that from Proposition 
\ref{prop:if_numerator},
\begin{align*}
    \sqrt{n}(\what{N}_Q-N_Q)&=
    \sum_{i=1}^n
    \psi_{N1}(\z_i)
    /\sqrt{n}
    +o_P(1)
\end{align*}
where $\psi_{N1}(\z)=g(\z)+h(\z)+\delta(\z)$ with
\begin{align*}
    g(\z) &= l\frac{t}{\pi^*(\featureval)}\lambda(\pi^*(\featureval),\alpha^*)-N_Q
\\
    h(\z)&=
 \E\left[
L\frac{T}{\pi^*(\feature)}\nabla_\alpha \lambda(\pi^*(\feature),\alpha^*)
\right] 
(t-\alpha^*)
\\
    \delta(\z) &= 
    \mu_Q(\featureval) \pi^*(\featureval) 
    (t-\pi^*(\featureval))
    \frac{\partial}{\partial \pi}\left[\frac{\lambda(\pi,\alpha^*)}{\pi}\right]\Big|_{\pi=\pi^*(\featureval)}
.\end{align*}
Note that we can also express $\delta(\z)$ as 
\begin{align*}
    \delta(\z) 
    &= \mu_Q(\featureval)(t-\pi^*(\featureval))\left(\nabla_\pi \lambda(\pi^*(\featureval),\alpha^*)-\frac{
        \lambda(\pi^*(\featureval),\alpha^*)}{\pi^*(\featureval)}\right)
    \\&=\mu_Q(\featureval)
    \left(
    \lambda(\pi^*(\featureval),\alpha^*)
    -\frac{t}{\pi^*(\featureval)}\lambda(\pi^*(\featureval),\alpha^*)
    +
(t-\pi^*(\featureval))
\nabla_\pi\lambda(\pi^*(\featureval),\alpha^*)
    \right)
    .
\end{align*}
It is clear that $\psi_{N1}(\Z)$ is the same 
as $\IF\{N_Q\}$ from the previous
section.
Then 
\begin{align*}
    \sqrt{n}(\what{N}_Q-N_Q) = \frac{1}{\sqrt{n}}\sum_{i=1}^n
    \psi_{N1}(\z_i)+o_P(1)\cd N(0,\var(\IF\{N_Q\}))
\end{align*}
where $\psi_{N1}(\z):=g(\z)+h(\z)+\delta(\z)$ as before.
An analogous argument applies to the denominator $D_Q$, 
as $D_Q$ is the same as $N_Q$ but with $L$ replaced by 1.
Then similar to before,
\begin{align*}
    \sqrt{n}(\what{D}_Q-D_Q) = \frac{1}{\sqrt{n}}\sum_{i=1}^n
    \psi_{D1}(\z_i)+o_P(1)\cd N(0,\var(\IF\{D_Q\}))
.\end{align*}
Then to see that
\begin{align*}
\sqrt{n}\left(\frac{\what{N}_Q}{\what{D}_Q}-\frac{N_Q}{D_Q}\right)
    =\frac{1}{\sqrt{n}}\sum_{i=1}^n \left(\frac{1}{D_Q}\psi_{N1}(\z_i) -
    \frac{N_Q}{D_Q^2}\psi_{D1}(\z_i)\right)
\cd N(0, \var(\IF\{N_Q/D_Q\}))
,\end{align*}
note that $\IF\{N_Q/D_Q\}$ is composed of $\IF\{N_Q\}$ and $\IF\{D_Q\}$
the same way that $\psi_{1}$ is composed of $\psi_{N1}$ and $\psi_{D_Q}$:
\begin{align*}
\IF\{N_Q/D_Q\}&=\frac{D_Q \IF\{N_Q\}-N_Q\IF\{D_Q\}}{D_Q^2}
=
[1/D_Q, -N_Q/D_Q^{2}]
\begin{bmatrix}
\IF\{N_Q\} \\
\IF\{D_Q\}
\end{bmatrix}
\end{align*}
which is the same as for 
$\hat\theta_Q=\what{N}_Q/\what{D}_Q$ as in Lemma~\ref{lem:fractions}:
\begin{align*}
\sqrt{n}(\what{N}_Q/\what{D}_Q-N_Q/D_Q)
&=[1/D_Q, -N_Q/D_Q^2]
\begin{bmatrix}
    \sum_{i=1}^n \psi_{N1}(\z_i)/\sqrt{n}+o_P(1)\\
    \sum_{i=1}^n \psi_{D1}(\z_i)/\sqrt{n}+o_P(1)
\end{bmatrix}
.\end{align*}
The results for the decomposition terms in Equation~\eqref{eq:decomp} follow
similarly.

%% file: more_estimation.tex
\section{Additional details for estimation algorithm}
\label{sec:more_estimation}
\subsection{Data splits and cross-fitting}
\label{sec:splits}
As is standard practice, performance 
for the model $f(\cdot)$ is not evaluated on training data
so that we measure loss degradation only from distribution shift,
rather than also from overfitting.
Thus, the data from training distribution $P$ has separate training and validation splits,
the evaluated model $f$ is trained 
on the training split, and evaluated on the validation split.
The evaluated model is also evaluated on the data from target distribution $Q$
(which consists of only one split). 
We measure the change in performance on $Q$ vs on the validation split of $P$. 

The practice of evaluating models on data separate
from the data they were trained on
also applies to the domain classifier
classifier $\hat\pi(\featureval)$. 
One natural option that uses all of the available data 
is called cross-fitting \cite{ZhengVa11,ChernozhukovChDeDuHaNeRo18,Yadlowsky22} 
in which 
the data are first split into $K$ folds.
Then, for each fold $k$, the domain classifier is learned on
the union of all the other folds, and finally evaluated on fold $k$.

There are various ways to set up the data splits. 
We use two different ways of splitting data in Sections \ref{sec:adult} and \ref{sec:wilds}
and describe them in \Appendix~\ref{sec:adult_splits} and
\ref{sec:wilds_splits}, respectively.

\subsection{Practical considerations for learning the domain classifier}
To estimate $\hat \pi(\featureval)=\P(T=1|\feature=\featureval)$,
we train classifier
on samples from $P_\featuresub$ and $Q_\featuresub$
using logistic loss, since the true $\pi(\featureval)$ minimizes the expected logistic loss,
and furthermore minimizers of the logistic loss produce values in $[0,1]$,
in contrast to squared loss.

To perform model checking, since 
the auxiliary domain classifier is analogous to a propensity score in
causal inference, the usual checks on propensity scores can be used to
check validity of the domain classifier.
These include checks for balance,
overlap, and calibration \cite{ImbensRu15,GutmanKaShi22}. 
It is also useful to check moments: for example, 
since $\E[\pi(\feature)]=\P(T=1)=\E[T]$, one should confirm that 
the sample mean of $\hat\pi(\feature)$ is also close to the sample mean of $T$.

%% file: experiment_details.tex
\section{Additional details for experiments}
\subsection{Adult dataset experiment details}
\label{appendix:adult_details}
\subsubsection{Models}
We use random forest classifiers using default settings from the
\texttt{sklearn} python package to fit both the employment model and the domain
classifier
$\hat \pi(x)$. The employment model has parameter \texttt{max\_depth} set to 2. 
\subsubsection{Data splits}
\label{sec:adult_splits}
Each of train and target datasets are split into an 80\%-20\% split.
The model to be evaluated is trained on the 80\% split of the train dataset.
The domain classifier is trained and the decomposition is evaluated on the 20\% split of the train dataset
and all of the target dataset, using 
cross-fitting with 3 splits as described 
in Section \ref{sec:splits}.
This way, both the model to be evaluated and the domain classifier are 
evaluated on data on which they are not trained. 

\subsubsection{Additional data processing}

The $X$ features are all of those in the Adult dataset unless otherwise
specified, but with some of them discretized and made into a binary encoding:
\begin{itemize}
    \item SCHL (educational attainment) was made into the following binary outcomes:
        \begin{itemize}
            \item Whether someone finished high school
            \item Whether someone finished college
            \item Whether someone finished a post-grad degree
        \end{itemize}
    \item MIL (military) was made into the following:
        \begin{itemize}
        \item Whether someone is active military
        \item Whether someone is a veteran
        \end{itemize}
    \item CIT (citizenship) was made into the following:
        \begin{itemize}
            \item Whether someone is born in the US
            \item Whether someone is born in a US territory
            \item Whether someone has American parents
            \item Whether someone is naturalized
            \item Whether someone is not a citizen
        \end{itemize}
    \item MIG (mobility) was made into the following:
        \begin{itemize}
            \item Whether someone moved residence
        \end{itemize}
    \item MAR (marriage) was made into the following:
        \begin{itemize}
            \item Whether someone is married
        \end{itemize}

\end{itemize}

\subsection{FMoW-wilds experiment details}
\label{appendix:wilds_details}

\subsubsection{Models}
The ERM and DANN models being evaluated are the ones trained
in the WILDS papers and downloaded from the WILDS leaderboard.
The domain classifiers are 
logistic regressions on top of last layer features from ERM and DANN, 
trained using \texttt{sklearn}~\cite{PedregosaVaGrMiThGrBlPr11}.
For both ERM and DANN, the WILDS leaderboard provides three models each
with different random seed, with the best hyperparameter settings.
We repeat the decomposition for each 
seed, and present the average of the results. 
Reported bootstrap standard errors are thus 
standard errors of this average. 

\subsubsection{Data splits}
\label{sec:wilds_splits}
The ERM and DANN models being evaluated 
have been trained on the \texttt{training} data split for ERM and DANN, and also on
\texttt{test-unlabeled} for DANN.
The domain classifiers are 
cross-fitted (with 2 splits) as described in Section
\ref{sec:splits} on \texttt{training} and
\texttt{test}. 
Then our decomposition is evaluated on \texttt{test} and the union of \texttt{id\_test} and
\texttt{id\_val}, which are separate from but drawn from the same distribution as
\texttt{training}. %
Because we used cross-fitting, we also evaluate the domain
classifier on a different cross-fitting split from which it was trained. 

Similar to \Appendix{} \ref{sec:adult_splits}, 
both the model to be evaluated and the domain classifier are 
evaluated on data on which they are not trained,
even though the data splits are different from those in \Appendix{}
\ref{sec:adult_splits}.

%% file: alternative_s.tex
\section{Alternative definitions of shared space}
\label{sec:alternative_s}
\begin{figure}[h]
        \centering
    \begin{subfigure}[b]{\linewidth}
        \centering
        \includegraphics[width=0.8\textwidth]{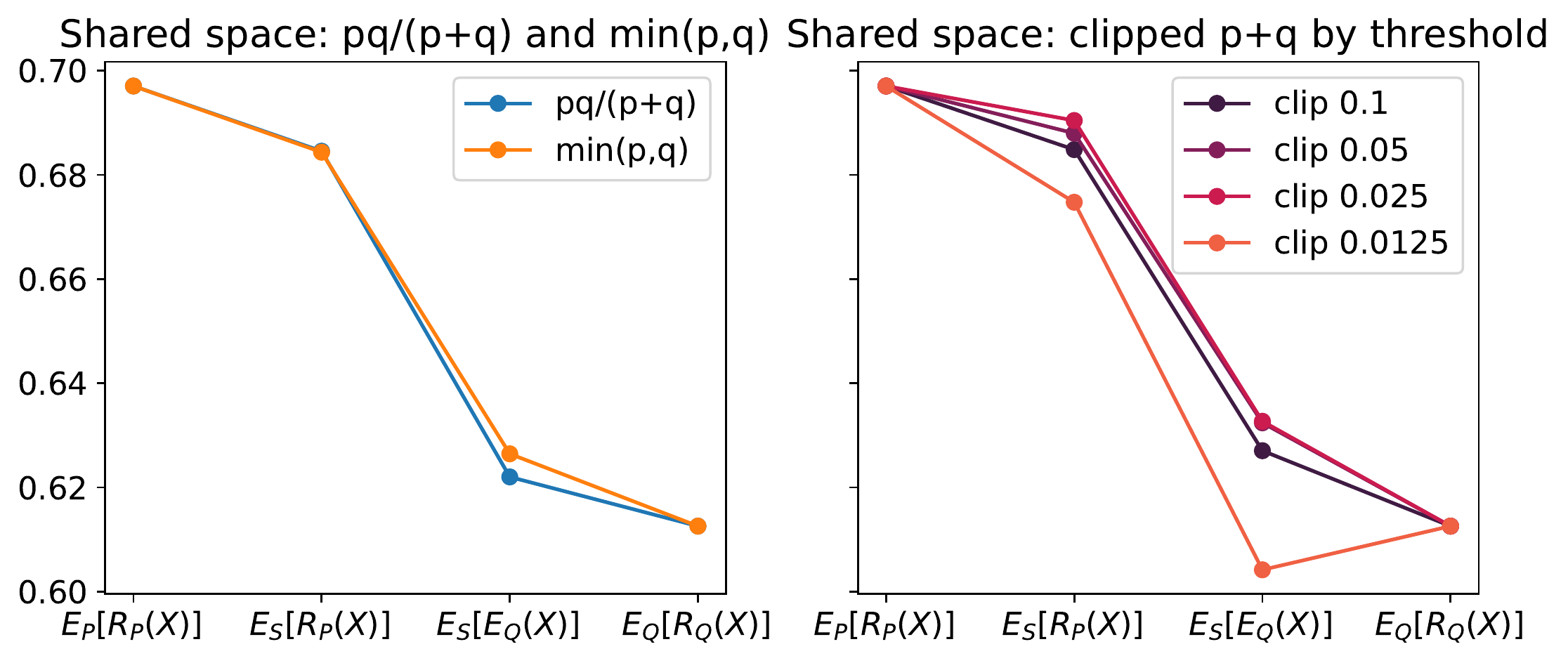}
        \caption{
        $Y|X$ shift: original model trained on West Virginia and evaluated on
        Maryland
        }
        \label{fig:alt_y_shift_missing}
    \end{subfigure}
    \begin{subfigure}[b]{\linewidth}
        \centering
        \includegraphics[width=0.8\linewidth]{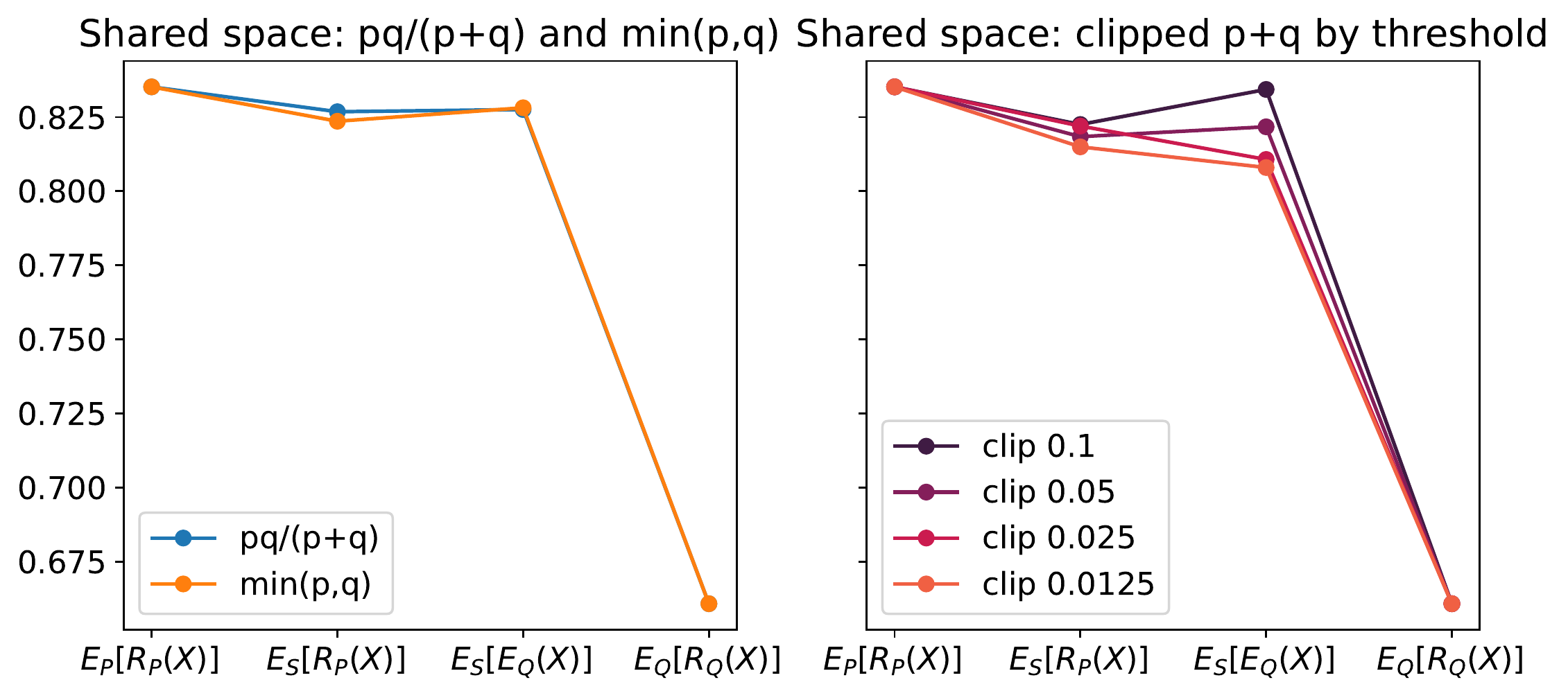}
        \caption{
$X$ shift: model trained on only age $\leq$25 and evaluated on general
        population
        }
        \label{fig:alt_x_shift_age_new}
    \end{subfigure}
    \begin{subfigure}[b]{\linewidth}
        \centering
        \includegraphics[width=0.8\linewidth]{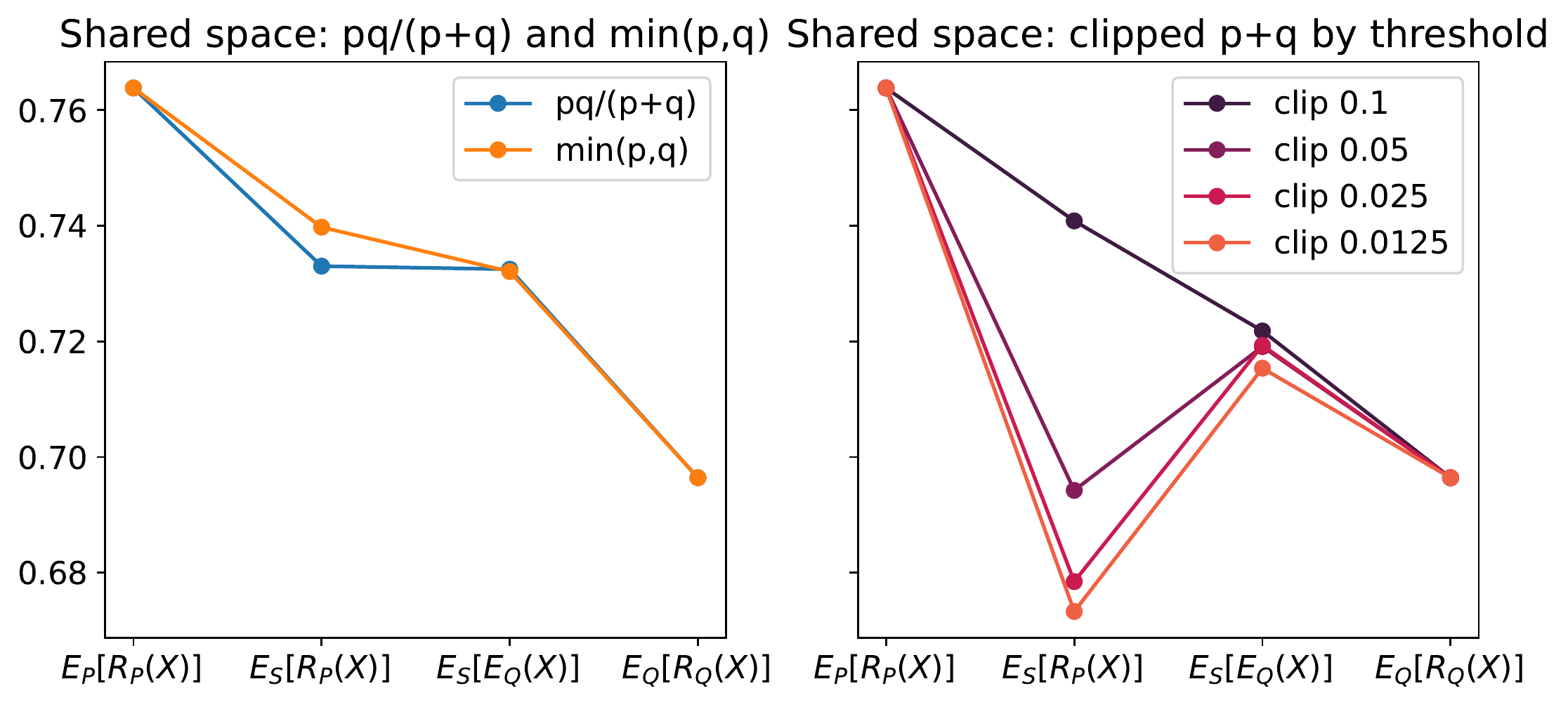}
        \caption{
            $X$ shift: model trained on over-sampling age $\leq 25$ and 
            evaluated on general population}
        \label{fig:alt_x_shift_age_oversample}
    \end{subfigure}

    \caption{Comparison of different shared spaces for experiments in
    Section~\ref{sec:adult}.
    On the $Y$ axis is accuracy; note that all results
    hold if we replace $\ell(\cdot)$ with accuracy. 
    The leftmost plots correspond to Equations~\eqref{eq:s} and
    \eqref{eq:s-minpq}, while the rightmost plots correspond
    to Equation~\eqref{eq:s-truncation}. 
    Observe that the decompositions tend to be qualitatively similar
    across different shared spaces, with the exception of
    those for
    Equation~\eqref{eq:s-truncation}. 
    Because of the need to choose a threshold for
    Equation~\eqref{eq:s-truncation} and the sensitivity of decompositions
    to the threshold value, we do not recommend it. 
    Note that the performance change attributed to $Y\mid X$ 
    does not always correspond to a \emph{worsening} of performance.
    In those cases, it could be noise, but it could also be
    that performance for $Q_{Y\mid X}$ is better than $P_{Y\mid X}$ on the
    shared $X$ distribution.
    }

    \label{fig:alt_decomps}
\end{figure}

The shared space $S_X$ we use in most of this work has density
\begin{align*}
s_X(x)\propto \frac{p_X(x)q_X(x)}{p_X(x)+q_X(x)}
\end{align*}
as first defined in Equation~\eqref{eq:s}, 
so that it has support only where both $p_X$ and $q_X$ has support,
and also higher (resp. lower) density when $p_X(x)$ and $q_X(x)$ have higher (resp. lower)
density. 
As mentioned in Equation \eqref{eq:alternative_s},
there are other definitions of $S_X$ that have similar properties. 
We repeat the experiments Section \ref{sec:adult}
with these alternative definitions of $S_X$ and show that the
decompositions are generally not too sensitive to 
the specific choice of $S_X$ in 
Figure~\ref{fig:alt_decomps}. 

